\documentclass[runningheads]{llncs}
\usepackage{graphicx}
\usepackage{comment}
\usepackage{amsmath,amssymb} 
\usepackage{color}

\usepackage[width=122mm,left=12mm,paperwidth=146mm,height=193mm,top=12mm,paperheight=217mm]{geometry}

\usepackage{amsmath,amssymb} 
\usepackage{hyperref}
\usepackage{tabularx}
\usepackage{soul}
\usepackage{makecell}
\usepackage{boldline}
\usepackage{subcaption}
\usepackage{booktabs}

\graphicspath{{assets/}} 
\newcolumntype{M}[1]{>{\centering\arraybackslash}m{#1}}

\newcommand{\norm}[1]{\left\lVert#1\right\rVert}
\newcommand{\abs}[1]{\left\lvert#1\right\rvert}

\begin{document}
\pagestyle{headings}
\mainmatter
\def\ECCVSubNumber{283}  

\title{DeepFit: 3D Surface Fitting via Neural Network Weighted Least Squares}

\titlerunning{DeepFit}
%
\author{Yizhak Ben-Shabat\inst{} \and
Stephen Gould\inst{}}
\authorrunning{Y. Ben-Shabat, S. Gould}
%
\institute{The Australian National University,
Australian Centre for Robotic Vision
\email{\{yizhak.benshabat,stephen.gould\}@anu.edu.au}}


\date{}

\maketitle
\begin{abstract}
We propose a surface fitting method for unstructured 3D point clouds. This method, called DeepFit, incorporates a neural network to learn point-wise weights for weighted least squares polynomial surface fitting. The learned weights act as a soft selection for the neighborhood of surface points thus avoiding the scale selection required of previous methods. To train the network we propose a novel surface consistency loss that improves point weight estimation. The method enables extracting normal vectors and other geometrical properties, such as principal curvatures, the latter were not presented as ground truth during training. We achieve state-of-the-art results on a benchmark normal and curvature estimation dataset, demonstrate robustness to noise, outliers and density variations, and show its application on noise removal.

\keywords{Normal estimation, surface fitting, least squares, unstructured 3D point clouds, 3D point cloud deep learning}
\end{abstract}

\section{Introduction}
\label{Sec:intro}

Commodity 3D sensors are rapidly becoming an integral component of autonomous systems. These sensors, e.g., RGB-D cameras or LiDAR, provide a 3D point cloud representing the geometry of the scanned objects and surroundings. This raw representation, however, is challenging to process since it lacks connectivity information or structure, and is often incomplete, noisy and contains point density variations. 
In particular, processing it by means of convolutional neural networks (CNNs)---highly effective for images---is problematic because CNNs require structured, grid-like data as input. 

When available,  additional local geometric information, such as the surface normal and principal curvatures at each point, induces a partial local structure and improves performance of different tasks for interpreting the scene, such as over-segmentation \cite{ben2018graph}, classification \cite{qi2017pointnet++} and surface reconstruction~\cite{guerrero2018pcpnet}.

Estimating the normals and curvatures from a raw point cloud with no additional information is a challenging task due to difficulties associated with sampling density, noise, outliers, and detail level. The common approach is to specify a neighborhood around a point  and  then fit a local basic geometric surface (e.g., a plane) to the points in this neighborhood. The normal at the point under consideration is estimated from the fitted geometric surface. The chosen size (or scale) of the neighborhood introduces an unavoidable trade-off between robustness to noise and accuracy of fine details. A large neighborhood over-smooths sharp corners and small details but is otherwise robust to noise. A small neighborhood, on the other hand, may reproduce the normals more accurately around small details but is more sensitive to noise. Evidently, a robust, scale-independent, data-driven surface fitting approach should improve normal estimation performance. 

We propose a surface fitting method for unstructured 3D point clouds. It features a neural network for point-wise weight prediction for weighted least squares fitting of polynomial surfaces. This approach removes the multi-scale requirement entirely and significantly increases robustness to different noise levels, outliers, and varying levels of detail. Moreover, the approach enables extracting normal vectors and additional geometric properties without the need for retraining or additional ground truth information. 
  The main contributions of this paper are: 
 \begin{itemize}
     \item A method for per-point weight estimation for weighted least squares fitting using deep neural networks.
     \item A scale-free method for robust surface fitting and normal estimation.
     \item A method for principal curvature and geometric properties estimation without using ground truth labels. 
 \end{itemize}
 
 \begin{figure*}[t]
\centering
	\includegraphics[width=0.98\linewidth]{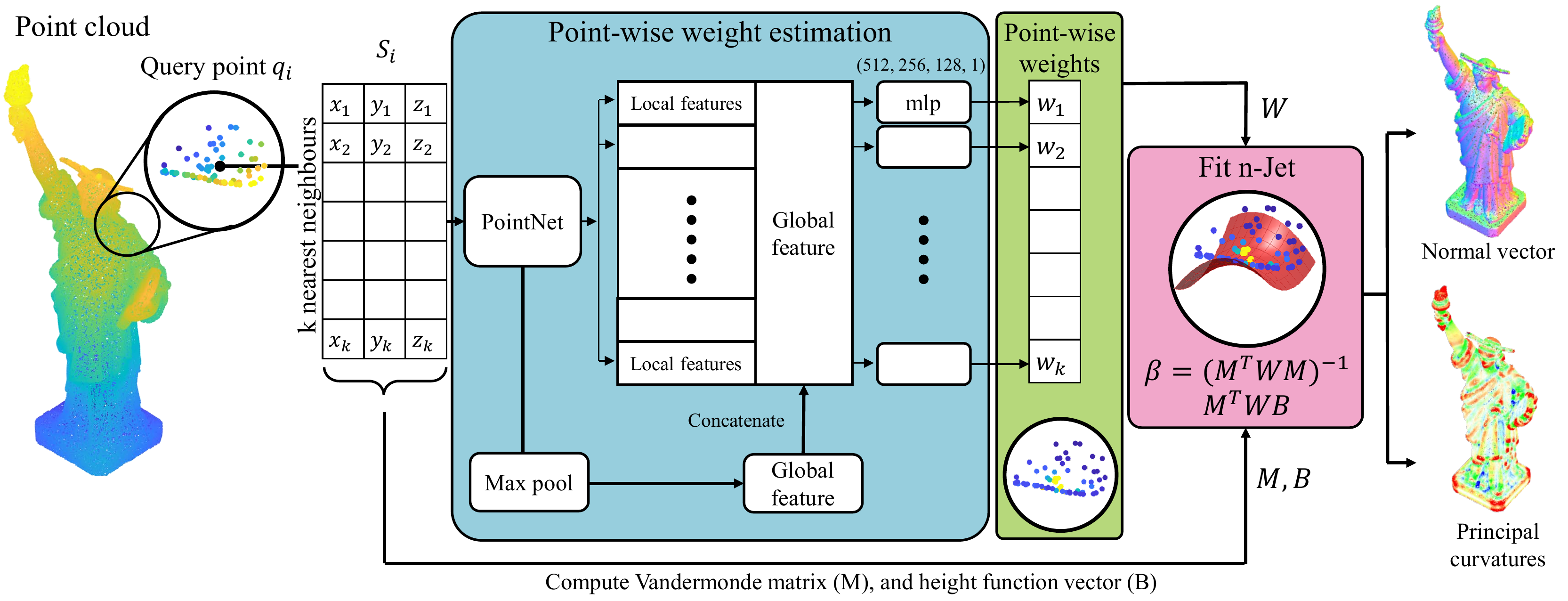}
	\caption{DeepFit pipeline for normal and principal curvature estimation. For each point in a given point cloud, we compute a global and local representation and estimate a point-wise wight. Then, we fit an $n$-jet by solving a weighted least squares problem.}
	\label{fig:approach} 
\end{figure*}
\section{Background and Related Work}
\label{Sec:related-work}

\subsection{Deep learning for unstructured 3D point clouds}
\label{SubSec:rel_work:DL_3D}
The point cloud representation of a 3D scene is challenging for deep learning methods because it is both unstructured and unordered. In addition, the number of points in the point cloud varies for different scenes. Several methods have been proposed to overcome these challenges. Voxel-based methods embed the point cloud into a voxel grid but suffer from several accuracy-complexity tradeoffs~\cite{maturana2015voxnet}. The PointNet approach \cite{Qi_2017_CVPR,qi2017pointnet++} applies a symmetric, order-insensitive, function on a high-dimensional representation of individual points. The Kd-Network \cite{klokov2017escape} imposes a kd-tree structure on the points and uses it to learn shared weights for nodes in the tree. The recently proposed 3D modified fisher vectors (3DmFV) \cite{ben20183dmfv} represents the points by their deviation from a Gaussian Mixture Model (GMM) whose Gaussians are uniformly positioned on a coarse grid. 

In this paper we use a PoinNet architecture for estimating point-wise weights for weighted least squares surface fitting. We chose PointNet since it operates directly on the point cloud, does not require preprocessing, representation conversion or structure, and contains a relatively low number of parameters, 

\subsection{Normal Estimation}
A classic method for estimating normals uses principal component analysis (PCA)~\cite{hoppe1992surface}. Here a neighborhood of points within some fixed scale is chosen and PCA regression used to estimate a tangent plane. Variants that fit local  spherical surfaces  \cite{guennebaud2007algebraic} or Jets \cite{cazals2005estimating} (truncated Taylor expansion) have also been proposed. Further detail on Jet fitting is given in Section \ref{subsec:jet_fitting_math}. To be robust to noise, these methods usually choose a large-scale neighborhood, leading them to smooth sharp features and fail to estimate normals near 3D edges. Computing the optimal neighborhood size can decrease the estimation error \cite{mitra2003estimating} but requires the (usually unknown) noise standard deviation value and a costly iterative process to estimate the local curvature and additional density parameters. 

A few deep learning approaches have been proposed to estimate normal vectors from unstructured point clouds. Boulch and Marlet proposed to transform local point cloud patches into a 2D Hough space accumulator by randomly selecting point triplets and voting for that plane's normal. Then, the normal is estimated from the accumulator by designing explicit criteria \cite{boulch2012fast} for bin selection or, more recently, by training a 2D CNN \cite{boulch2016deep} to estimate it continuously as a regression problem. This method does not fully utilize available 3D information since it loses information during the transformation stage. Another method, named PCPNnet \cite{guerrero2018pcpnet}, uses a PointNet \cite{Qi_2017_CVPR} architecture over local neighborhoods at multiple scales. It achieves good normal estimation performance and has been extended to estimating principal curvatures. However, it processes the multi-scale point clouds jointly and requires selecting a predefined set of scales.  
A more recent work, Nesti-Net \cite{ben2019nesti} tries to predict the appropriate scale using a mixture of experts network and a local representation for different scales. It achieves high accuracy but suffers from high computation time due to the multiple scale computations. Nesti-Net shares PCPNet's drawback of requiring a predefined set of scales. 
A contemporary work \cite{lenssen2019differentiable} uses an iterative plane fitting approach which tries to predict all normals of a local neighborhood and iteratively adjusts the point weights to best fit the plane. 

In this paper we propose a novel approach for normal estimation by learning to fit an $n$-order Jet while predicting informative points' weights. Our approach removes the need of predefined scales and optimal scale selection since the informative points are extracted at any given scale. Our method generalizes the contemporary method proposed by Lenssen et. al.~\cite{lenssen2019differentiable}, avoids the iterative process, and enables the computation of additional geometric properties.

\subsection{Jet fitting using least squares and weighted least squares}
\label{subsec:jet_fitting_math}
We now provide background and mathematical notation for truncated Taylor expansion surface fitting using least-squares (LS) and weighted least-squares (WLS). We refer the interested reader to Cazals and Pouget~\cite{cazals2005estimating} for further detail. 

Any regular embedded smooth surface can be locally written as the graph of a bi-variate ``height function'' with respect to any z-direction that does not belong to the tangent space~\cite{spivak1970comprehensive}. We adopt the naming convention of Cazals and Pouget~\cite{cazals2005estimating} and refer to the truncated Taylor expansion as a degree $n$ jet or $n$-jet for short. An $n$-jet of the height function over a surface is given by:

\begin{equation}
\label{eq:jet}
    f(x,y)=J_{\beta,n}(x,y)= \sum_{k=0}^{n}\sum_{j=0}^{k}\beta_{k-j,j}x^{k-j}y^j
\end{equation}
\\
Here $\beta$ is the jet coefficients vector that consists of $N_n=(n+1)(n+2)/2$ terms.

In this work we wish to fit a surface to a set of $N_p$ 3D points. For clarity, we move to the matrix notation and specify the Vandermonde matrix $M=(1, x_i, y_i, ..., x_i y_i^{n-1}, y_i^n)_{i=1,...,N_p}\in \mathbb{R}^{N_p \times N_n}$ and the height function vector $B=(z_1, z_2,...z_{N_p})^T \in \mathbb{R}^N_p$ representing the sampled points. We require that every point satisfy Eq. \ref{eq:jet}, yielding the system of linear equations:
\begin{equation}
    M\beta = B
\end{equation}
When $ N_n > N_p$ the system is over-determined and an exact solution may not exist. Therefore we use an LS approximation that minimizes the sum of square errors between the value of the jet and the height function over all points:
\begin{equation}
    \beta = \arg\min_{z \in \mathbb{R}^{N_n}} \norm{Mz - B}^2
\end{equation}
It is well known that the solution can be expressed in closed-form as: 
\begin{equation}
\label{eq:lstsqr_sol}
    \beta = (M^TM)^{-1}M^TB
\end{equation}
\\
Typically the sampled points include noise and outliers that heavily reduce the fitting accuracy. To overcome this, the formulation given in Eq.~\ref{eq:lstsqr_sol} can be extended to a weighted least square problem. In this setting, some points have more influence on the fitted model than others. Let $W\in\mathbb{R}^{N_p \times N_p}$ be a diagonal weight matrix $W=\textrm{diag}(w_1, w_2, ..., w_{N_p})$. Each element in the matrix's diagonal $w_i$ corresponds to the weight of that point. 
\\
The optimization problem becomes: 
\begin{align*}
    \beta &= \arg\min_{z \in \mathbb{R}^{N_n}} \norm{W^{1/2}(Mz - B)}^2\\
    & = \arg\min_{z \in \mathbb{R}^{N_n}}\sum_{i=1}^{N_p} w_i \left(\sum_{j=1}^{N_n} M_{ij}z_j - B_i\right)^2
\end{align*}
and its solution:
\begin{equation}
\label{eq:wlstsqr_sol}
    \beta = (M^TWM)^{-1}M^TWB
\end{equation}

In this work, we choose to focus on $n$-jet fitting because any order $n$ differential quantity can be computed from the $n$-jet. This is one of the main advantages of our method. That is, our method is trained for estimating normal vectors but is then able to estimate other differential quantities, e.g., principal curvatures, depending on the jet order.
\section{DeepFit}
\label{Sec:DeepFit}

\subsection{Learning point-wise weights}
\label{subSec:learning_weights}

The full pipeline for our method is illustrated in Fig.~\ref{fig:approach}.
Given a 3D point cloud $S$ and a query point $q_i \in S$ we first extract a local subset of points $S_i$ using k-nearest neighbors.  We then use a neural network to estimate the weight of each point in the neighborhood, which will subsequently be used for weighted least squares surface fitting.
Specifically, we feed $S_i$ into a PointNet~\cite{Qi_2017_CVPR} network, which outputs a global point cloud representation $G(S_i)$. Additionally, we extract local representations from an intermediate layer for each of the points $p_{j} \in S_i$ separately to give $g(p_{j})$. These representations are then concatenated and fed into a multi-layer perceptron $h(\cdot)$ followed by a sigmoid activation function. We choose a sigmoid in order to limit the output values to be between 0 and 1. The output of this network is a weight per point that is used to construct the diagonal point-weight matrix, $W = \text{diag}(w_j)$ with
\begin{align}
w_j &= \text{sigmoid}(h(G(S_i), g(p_{ij}))) + \epsilon
\end{align}
\\
For numerical stability, we add a constant small $\epsilon$ in order to avoid the degenrate case of a zero or poorly conditioned matrix. This weight matrix is then used to solve the WLS problem of Eq.~\ref{eq:wlstsqr_sol} and approximate the $n$-jet coefficients $\beta$. All parts of the network are differentiable and therefore it is trained end-to-end. 

\subsection{Geometric quantities estimation}
\label{subSec:approach:normal_estimation}
Given the $n$-jet coefficients $\beta$ several geometric quantities can be easily extracted:
\subsubsection{Normal estimation.}
The estimated normal vector is given by:
\begin{equation}
    N_i=\frac{(-\beta_{1}, -\beta_{2}, 1)}{\norm{(-\beta_{1}, -\beta_{2}, 1)}_2}
\end{equation}

\subsubsection{Shape operator and principal curvatures.}
For the second order information we compute the Weingarten map of the surface by multiplying the inverse of the first fundamental form and the second fundamental form. Its eigenvalues are the principal curvatures $(k_1, k_2)$, and its eigenvectors are the principal directions. The computation is done in the tangent space associated with the parametrization.  
\begin{align}
    M_{\text{Weingarten}} &= -\frac{1}{\sqrt{\beta_1^2+\beta_2^2+1}}
    \begin{bmatrix} 
    1+\beta_1^2 & \beta_1\beta_2 \\
    \beta_1\beta_2 & 1+\beta_2^2
    \end{bmatrix}^{-1}\begin{bmatrix} 
    2\beta_3 & \beta_4 \\
    \beta_4 & 2\beta_5
    \end{bmatrix}
\end{align}
Generally, the principal curvatures can be used as ground truth in training, however, due to the eigenvalue decomposition, with the high probability of outputting two zero principal curvatures (planes) it suffers from numerical issues when computing the gradients for backpropagation \cite{dang2018eigendecomposition}. Therefore, we compute the curvatures only at test time. Note that Monge basis and higher order Monge coefficients can also be computed, similar to \cite{cazals2005estimating}.


\subsection{Consistency loss}
\label{subSec:con_loss}
In order to learn point-wise weights, we introduce a local consistency loss $L_{con}$. This loss is composed of two terms, the weighted normal difference term and a regularization term. The weighted normal difference term computes a weighted average of the sine of the angle  between the ground truth normal and the estimated normal at every local neighborhood point. These normals are computed analytically by converting the $n$-jet to the implicit surface form of $F(x,y,z)=0$. Therefore, for every query point $q_i$ and its local neighborhood $S_i$ we can compute the normal at each neighboring point $p_{j} \in S_i$ using: 
\begin{equation}
    N_{j} = \left.\frac{\nabla F}{\norm{\nabla F}}\right|_{p_{j}} =
     \left.\frac{(-\beta_i \frac{\partial M}{\partial x}^T,
     \beta_i \frac{ \partial M}{\partial y}^T, 1)}{\norm{\nabla F}}\right|_{p_{j}}
\end{equation}
Note that this formulation assumes all points to lie on the surface, for points that are not on the surface, the normal error will be large, therefore that points weight will be encouraged to be small. This term can easily converge to an undesired local minimum by setting all weights to zero. In order to avoid that, we add a regularization term which computes the negative average $log$  of all weights. 
In summary, the consistency loss for a query point $q_i$ is then given by:
\begin{equation}
    L_{con} =\frac{1}{N_{q_i}} \left[ -\sum_{j=1}^{N_{q_i}}log(w_{j}) + \sum_{j=1}^{N_{q_i}}w_{j} \abs{N_{GT}\times N_{j}} \right]
\end{equation}

In contrast to Lenssen et. al. \cite{lenssen2019differentiable}, this formulation allows us to avoid solving multiple linear systems iteratively for each point in the local neighborhood.

In total, to train the network, we sum several loss terms: The $sin$ loss between the estimated unoriented normal and the ground truth normal at the query point, the consistency loss, and PointNet's transformation matrix regularization terms $L_{reg}=\abs{I-AA^T}$. 

\begin{equation}
    L_{tot} = \abs{N_{GT}\times N_{i}} + \alpha_1L_{con} +\alpha_2L_{reg}
\end{equation}
\\
Here, $\alpha_1$, and $\alpha_2$ are weighting factors, chosen empirically.

\subsection{Implementation notes}
\label{subSec:impl_notes}

In our experiments we report results using DeepFit with the following configuration, unless otherwise stated. A four layer MLP with sizes 512, 256, 128, and 1; a neighborhood size of 256 points, and a 3-order jet.
In order to avoid numerical issues, simplify the notation, and reduce the linear algebra operations,  we perform the following pre-processing stages on every local point cloud:
\begin{enumerate}
    \item Normalization: we translate the point cloud to position the query point in the origin and scale the point cloud to fit a unit sphere. 
    \item Basis extraction: we perform principal component analysis (PCA) on the point cloud. We then use the resulting three orthonormal eigenvectors as the fitting basis so that the vector associated with the smallest eigenvalue is the last vector of the basis. 
    \item Coordinate frame transformation: We perform a change of coordinates to move the points into the coordinate system of the fitting basis.
    \item Preconditioning: we precondition the Vandermonde matrix by performing column scaling. Each monomial $x_i^ky_i^l$ is divided by $h^{k+l}$. That is, $M' = MD^{-1}$ with $D$ the diagonal matrix $D=\text{diag}(1, h, h^2, ..., h^n)$. We use the mean of the norm $\norm{(x_i, y_i)}$ as $h$. The new system is then $M'(D\beta)=B$ and $\beta=D^{-1}(M'^TWM')^{-1}M'^TWB$.
\end{enumerate}

\noindent
Note that after the normal is estimated we apply the inverse transform to output the result in the original coordinate frame. 
\section{Results}
\label{Sec:results}

\subsection{Dataset and training details} 
For training and testing we used the PCPNet shape dataset \cite{guerrero2018pcpnet}. The training set consists of eight shapes: four CAD objects (fandisk, boxunion, flower, cup) and four high quality scans of figurines (bunny, armadillo, dragon and turtle). All shapes are given as triangle meshes and densely sampled with 100k points. The data is augmented by introducing i.i.d. Gaussian noise for each point's spacial location with a standard deviation of  0.012, 0.006, 0.00125 w.r.t the bounding box size. This yields a set with 3.2M training examples. The test set consists of 22 shapes, including figurines, CAD objects, and analytic shapes. For evaluation we use the same 5000 point subset per shape as in Guerrero et al.~\cite{guerrero2018pcpnet}.

All variations of our method were trained using 32,768 (1024 samples by 32 shapes) random subsets of the 3.2M training samples at each epoch. We used a batch size of $256$, the Adam optimizer and a learning rate of $10^{-3}$. The implementation was done in PyTorch and trained on a single Nvidia RTX 2080 GPU.

\subsection{Normal estimation performance}
\label{SubSec:results:baseline_n_est}

 We use the RMSE metric for comparing the proposed DeepFit to other deep learning based methods \cite{guerrero2018pcpnet,ben2019nesti,lenssen2019differentiable} and classical geometric methods \cite{hoppe1992surface,cazals2005estimating}. 
 Additionally, we analyze robustness for two types of data corruption: 
\begin{itemize}
    \item  Point density---applying two sampling regimes for point subset selection: gradient, simulating effects of distance from the sensor, and stripes, simulating local occlusions.    
    \item Point perturbations--adding Gaussian noise to the points coordinates with three levels of magnitude specified by $\sigma$, given as a percentage of the bounding box.
\end{itemize}
For the geometric methods, we show results for three different scales: small, medium and large, which correspond to 18, 112, 450 nearest neighbors. For the deep learning based methods we show the results for the single-scale (ss) and multi-scale (ms) versions.

Table \ref{table:results:baselines} shows the unoriented normal RMSE results for the methods detailed above. It can be seen that our method slightly outperforms all other methods for low, medium and no noise augmentation and for gradient density augmentation. For high noise, and striped occlusion augmentation we are a close second to the contemporary work  of Lenssen et al. \cite{lenssen2019differentiable} which only estimates the normal vectors while DeepFit also estimates other geometric properties, e.g., principal curvatures. 
The results also show that all method's performance deteriorate as the noise level rises. In this context, both PCA and Jet perform well for specific noise-scale pairs. In addition, for PCPNet, using a multiple scales only mildly improves performance. Nesti-Net's mixture of experts mitigate the scale-accuracy tradeoff well at the cost of computational complexity. DeepFit's soft point selection process overcomes this tradeoff. In the supplemental materials we perform additional evaluation using the percentage of good points (PGP$\alpha$) metric.

\begin{table*} 
	\centering	
		\tabcolsep = 0.0025\textwidth
		\begin{tabular}{| M{0.12\textwidth} | M{0.10\textwidth} | M{0.06\textwidth} | M{0.06\textwidth}| M{0.06\textwidth} | M{0.06\textwidth}|
		M{0.06\textwidth}| M{0.06\textwidth} | M{0.06\textwidth}|  M{0.07\textwidth} | M{0.08\textwidth} | M{0.12\textwidth} |} 
			\hline
			\centering\textbf{Aug.}  & \centering\textbf{Our DeepFit}  & \multicolumn{3}{c|}{\makecell[{{M{0.18\textwidth}}}]{\textbf{PCA} \\ \cite{hoppe1992surface}} }  &
			\multicolumn{3}{c|}{\makecell[{{M{0.18\textwidth}}}]{\textbf{Jet} \\ \cite{cazals2005estimating}} } & 	\multicolumn{2}{c|}{\makecell[{{M{0.14\textwidth}}}]{\textbf{PCPNet} \\ \cite{guerrero2018pcpnet}} } & \centering\textbf{Len-ssen et. al} \\ \cite{lenssen2019differentiable} & \centering\textbf{Nesti-Net}
 			\tabularnewline
 			\hline
 			scale &ss&small&med&large&small&med&large&ss&ms&ss&ms (MoE)\\
 			\hlineB{2}
 		    None               &\textbf{6.51}&8.31&12.29&16.77&7.60&12.35&17.35&9.68&9.62&6.72&6.99\\
 		    \hline
 			\textbf{Noise $\sigma$}     &&&&&&&&&&& \\
 			$0.00125$ &\textbf{9.21}&12.00&12.87&16.87&12.36&12.84&17.42&11.46&11.37&9.95&10.11\\
 			$0.006$   &\textbf{16.72}&40.36&18.38&18.94&41.39&18.33&18.85&18.26&18.87&17.18&17.63\\
 			$0.012$   &23.12&52.63&27.5 &23.5 &53.21&27.68&23.41&22.8&23.28&\textbf{21.96}&22.28\\
		    \hline
		    \textbf{Density}   &&&&&&&&&&& \\
		    Gradient           &\textbf{7.31}&9.14&12.81&17.26&8.49&13.13&17.8&13.42&11.7 &7.73&9.00\\
		    Stripes            &7.92&9.42&13.66&19.87&8.61&13.39&19.29&11.74&11.16&\textbf{7.51}&8.47\\
		    \hline
		    \textbf{average}   &\textbf{11.8}&21.97&16.25&18.87&21.95&16.29&19.02&14.56&14.34&11.84&12.41 \\
		    \hline
		\end{tabular}
	\caption{Comparison of the RMSE angle error for unoriented normal vector estimation of our DeepFit method to classical geometric methods (PCA \cite{hoppe1992surface} and Jet \cite{cazals2005estimating} - for three scales small, med, and large corresponding to $k=18, 122, 450$), and deep learning methods (PCPNet \cite{guerrero2018pcpnet}, Lenssen et. al \cite{lenssen2019differentiable}, and Nesti-Net \cite{ben2019nesti})}
	\label{table:results:baselines}
\end{table*}

\begin{figure}
\centering
    \begin{subfigure}{.48\textwidth}
        \centering
    	\includegraphics[width=0.98\linewidth]{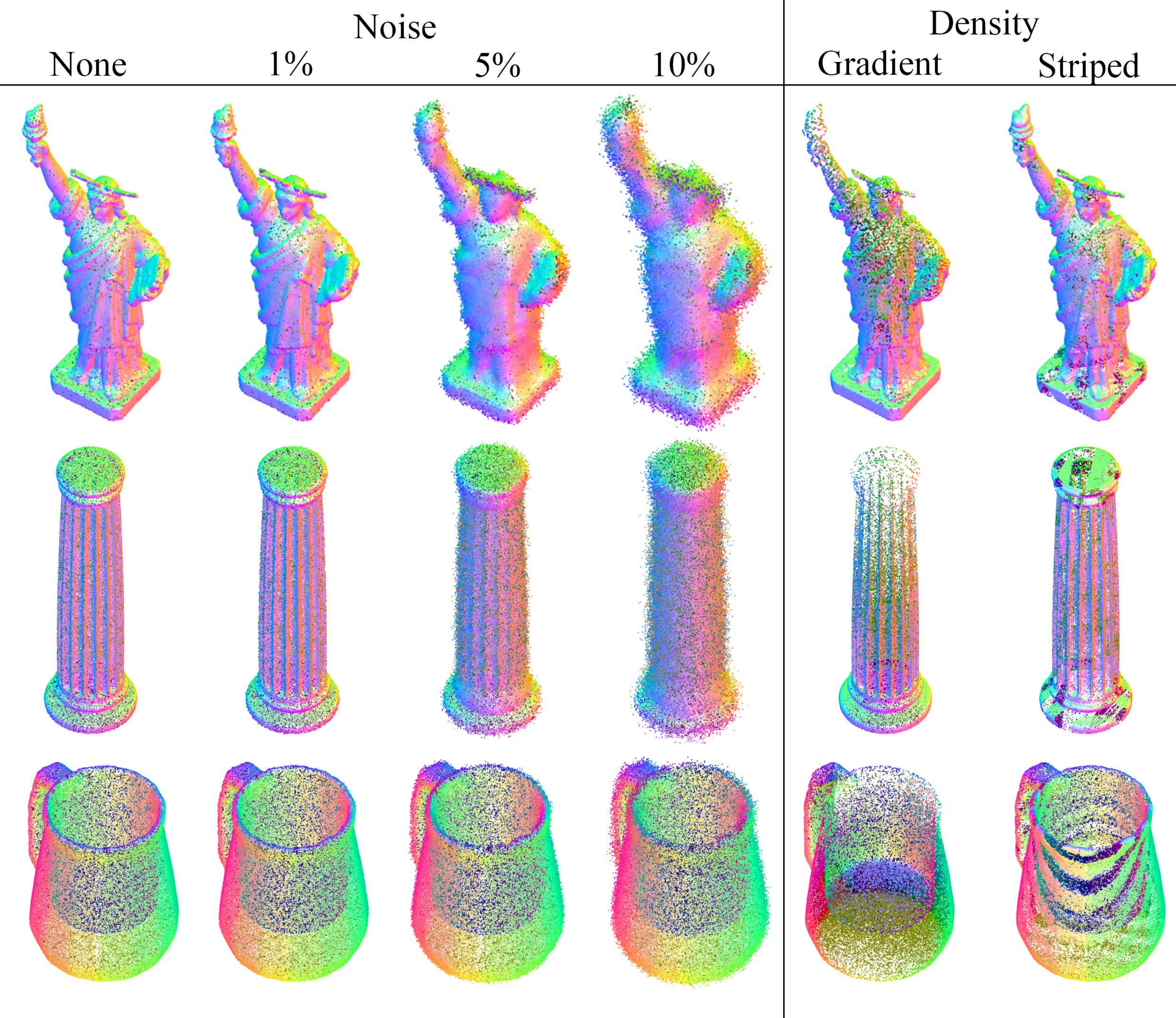}
    	\caption{}
    	\label{fig:results_normals_visualiztion:a}
    \end{subfigure}
    \unskip\ \vrule\ 
    \begin{subfigure}{.48\textwidth}
        \centering
    	\includegraphics[width=0.98\linewidth]{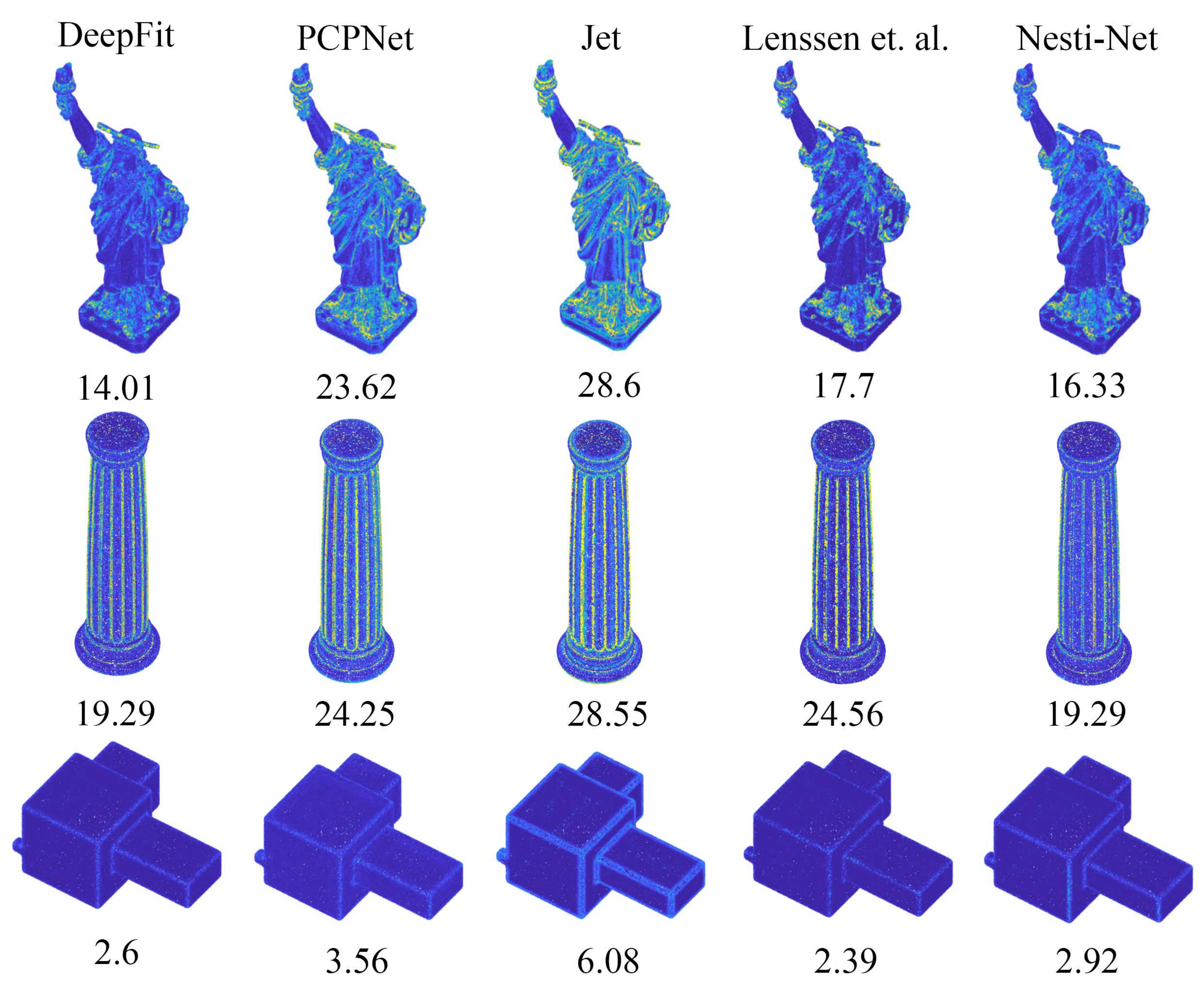}
        	\caption{}
        	\label{fig:results_normals_visualiztion:b}
	\end{subfigure}
	\caption{(a) DeepFit's normal estimation results for different noise levels (columns 1-4), and density distortions (columns 5-6). The colors of the points are normal vectors mapped to RGB. (b) Normal estimation error visualization results of DeepFit compared to other methods for three types of point clouds without noise. The colors of the points correspond to angular difference, mapped to a heatmap ranging from 0-60 degrees.}
\end{figure}

Figure \ref{fig:results_normals_visualiztion:a} depicts a visualization of DeepFit's results on three point clouds. Here the normal vectors are mapped to the RGB cube. 
It shows that for complex shapes (pillar, liberty) with high noise levels, the  general direction of the normal vector is predicted correctly, but, the fine details and exact normal vector are not obtained. For a basic shape (Boxysmooth) the added noise does not affect the results substantially. Most notably, DeepFit shows robustness to point density corruptions.
Figure \ref{fig:results_normals_visualiztion:b} depicts a visualization of the angular error in each point for the different methods using a heat map. For the Jet method~\cite{cazals2005estimating} we display the results for medium scale. For all methods, it can be seen that more errors occur in regions with small details, high curvature e.g. edges and corners, and complex geometry. DeepFit suffers the least from this effect due to its point-wise weight estimation, which allows it to adapt to the different local geometryand disregard irrelevant points in the fitting process. 

\begin{figure}
\centering
	\includegraphics[width=0.98\linewidth]{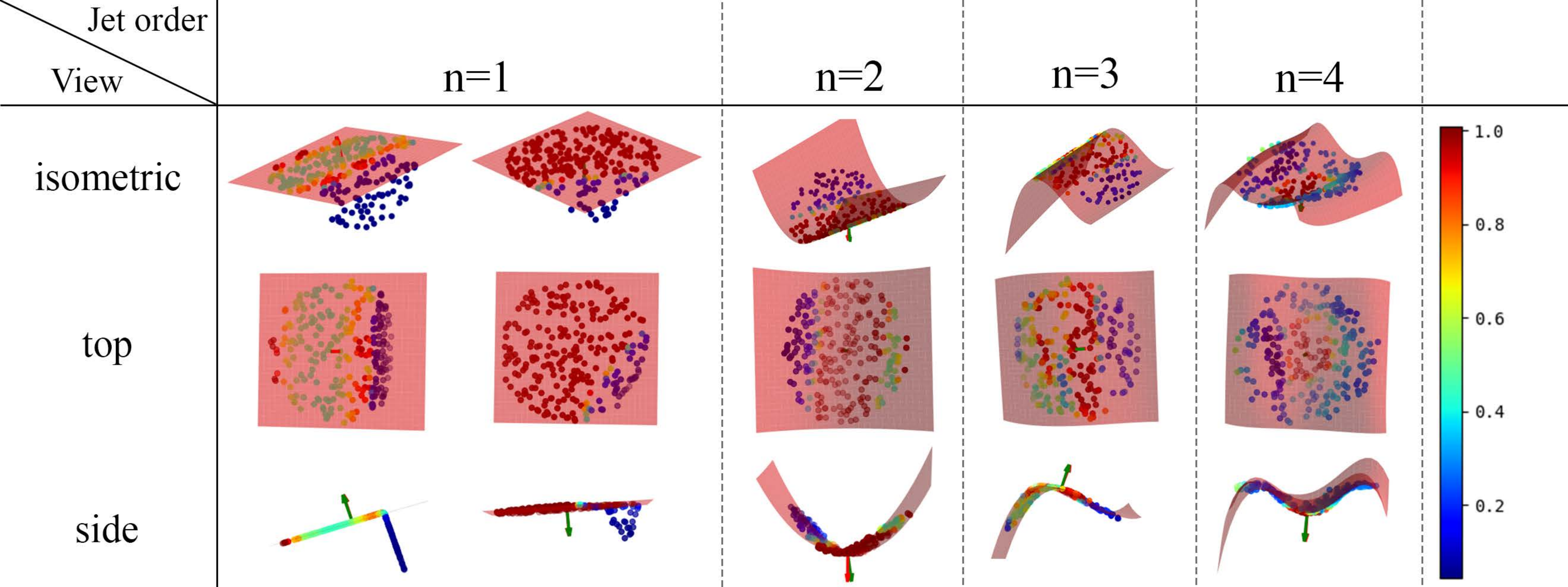}
	\caption{DeepFit point-wise weight prediction. Three views of different $n$-jet surface fits. The colors of the points correspond to weight magnitude , mapped to a heatmap ranging from 0 to 1; see color bar on the right i.e. red points highly affect the fit while blue points have low influence.}.
	\label{figure:results_weights_visualiztion}
\end{figure}

Figure \ref{figure:results_weights_visualiztion} qualitatively visualizes the performance of DeepFit's point-wise weight prediction network. The colors of the points correspond to weight magnitude, mapped to a heatmap ranging from 0 to 1 i.e. red points highly affect the fit while blue points have low influence. It shows that the network learns to adapt well to corner regions (column $n=1$), assigning high weights to points on one plane and excluding points on the perpendicular one. Additionally, it shows how the network adapted the weight to achieve a good fit for complex geometries (column $n=2,3,4$). 

\begin{figure}[t]
    \centering
    	\begin{subfigure}{.98\textwidth}
    \centering
\includegraphics[width=0.4\linewidth]{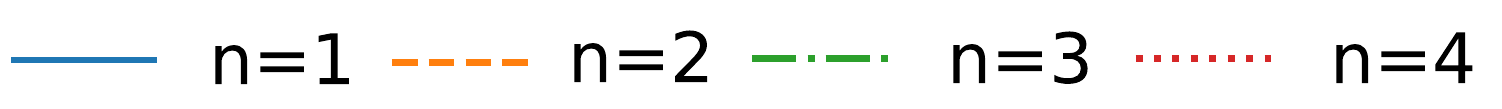}
    	\label{fig:results:ablations:b}
	\end{subfigure}

    \begin{subfigure}{.49\textwidth}
        \centering
    	\includegraphics[width=0.98\linewidth]{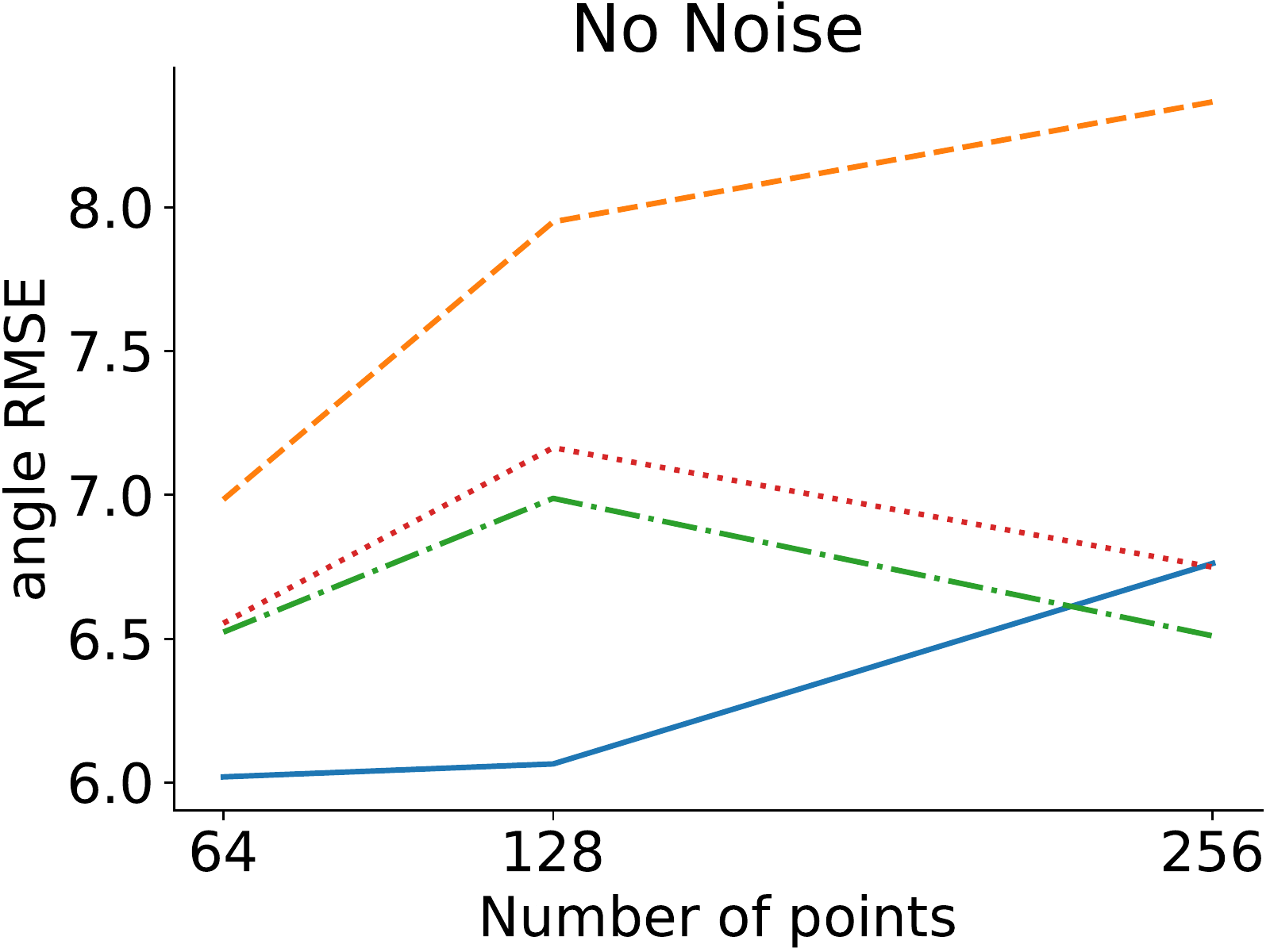}
    	\caption{}
        \label{fig:results:ablations:a}
    \end{subfigure}
	\begin{subfigure}{.49\textwidth}
    \centering
\includegraphics[width=0.98\linewidth]{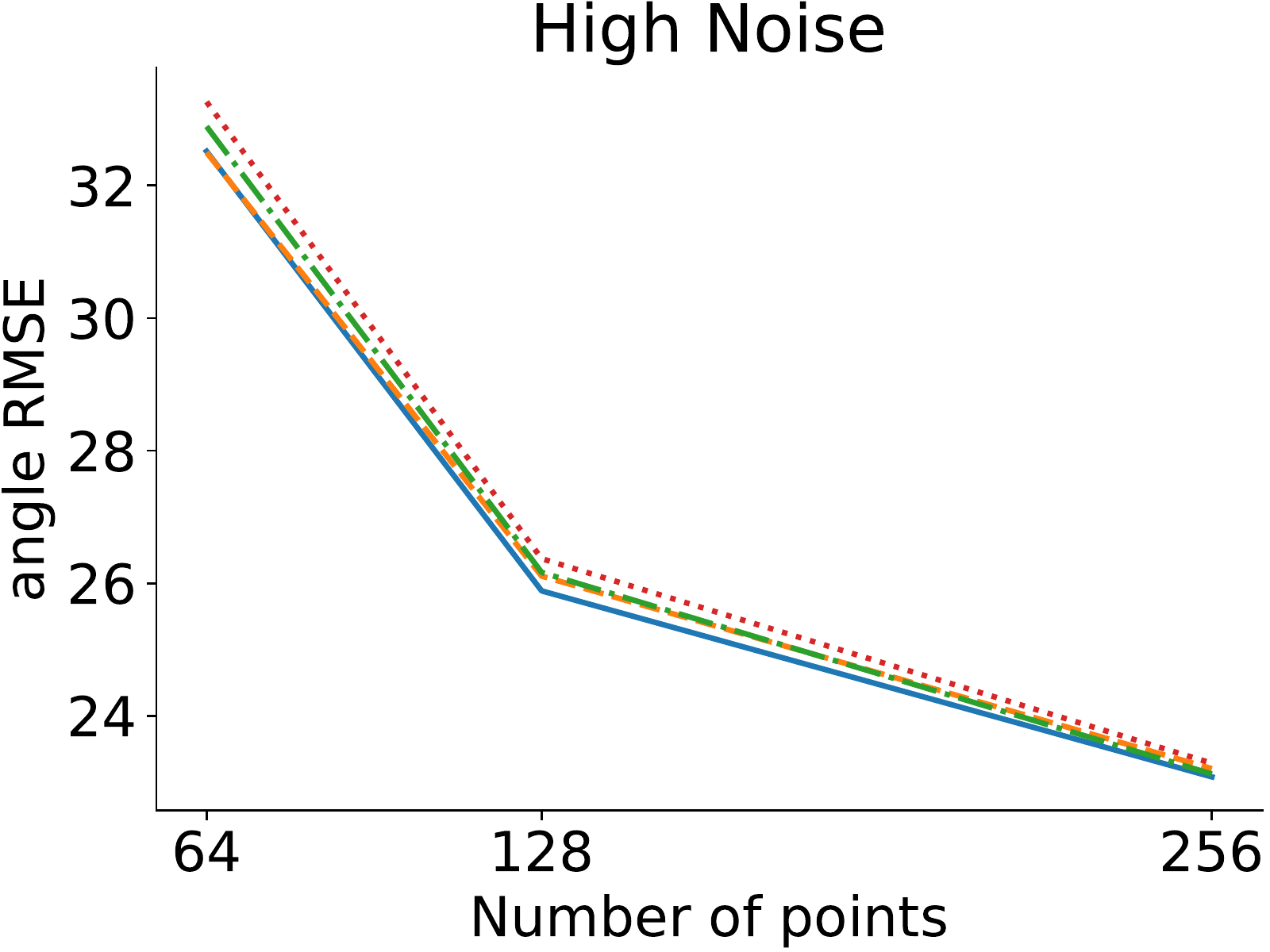}
    	\caption{}
    	\label{fig:results:ablations:b}
	\end{subfigure}
	    \caption{Normal estimation RMSE results for DeepFit ablations for (a) no noise and (b) high noise augmentations. Comparing the effect of number of neighboring points and jet order.}
    
    \label{fig:results:ablation}
\end{figure}

Fig. \ref{fig:results:ablation} shows the unoriented normal RMSE results for different parameter choices of our method. We explore different Jet orders $n=1, 2, 3, 4$, and a different number of neighboring points $k=64, 128, 256$, It shows that using a large neighborhood size highly improves the performance in high noise cases while only minimally affecting the performance in low noise. It also shows that all jet orders are comparable with a small advantage for order 1-jet (plane) and order 3-jet which is an indication for a bias in the dataset towards low curvature geometry. Additional ablation results, including more augmentations and the PGP$\alpha$ metric are provided in the supplemental material.

Timing and efficiency performance are provided in the supplemental material. DeepFit is faster and has fewer parameters than PCPNet and Nesti-Net and has the potential of only being slightly slower than CGAL implementation of Jet fitting because the forward pass for weight estimation is linear with respect to the number of points and the network weights.

\subsection{Principal curvature estimation performance}
\label{SubSec:results:baseline_c_est}
\begin{figure}
\centering
    	\includegraphics[width=0.80\linewidth]{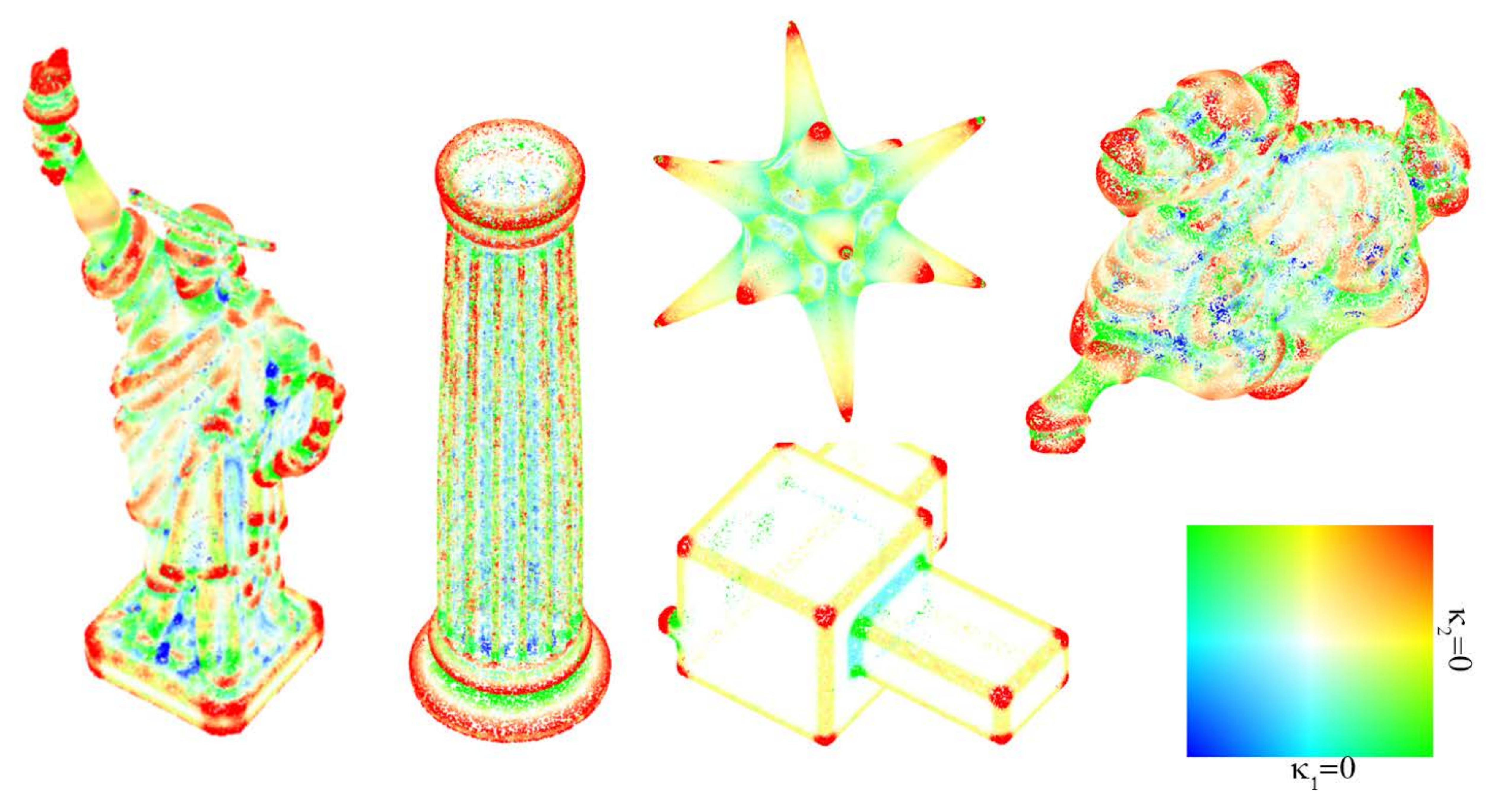}
	\caption{Curvature estimation results visualization.  The colors of the points corresponds to the mapping of $k_1, k_2$  to the color map given in the bottom right. Values in the range $[-(\mu(|k_i|)+\sigma(|k_i|)),\mu(|k_i|)+\sigma(|k_i|)] |_{i=1,2}$.}
	\label{fig:curvature_result_visualization}
\end{figure}

Figure \ref{fig:curvature_result_visualization} qualitatively depicts DeepFit's results on five point clouds. For visualization, the principal curvatures are mapped to RGB values according to the commonly used mapping given in its bottom right corner i.e. both positive (dome) are red, both negative (bowl) are blue, one positive and one negative (saddle) are green, both zero (plane) are white, and one zero and one positive/negative (cylinder) are yellow/cyan. For consistency in color saturation we map each model differently according to the mean and standard deviation of the principal curvatures.  Note that the curvature sign is determined by the ground truth normal orientation. 

For quantitative evaluation we use the normalized RMSE metric curvature estimation evaluation proposed in Guerrero et. al.~\cite{guerrero2018pcpnet} and given in Eq. \ref{eq:D_k}, for comparing the proposed method to other deep learning based  \cite{guerrero2018pcpnet} and geometric methods \cite{cazals2005estimating}. Table \ref{table:results:curvature_baselines} summarizes the results and shows an average error reduction of 35\% and 13.7\% for maximum and minimum curvatures respectively. We analyze robustness for the same types of data corruptions as in normal estimation i.e. point perturbation and density. DeepFit significantly outperforms all other methods for maximum principal curvature $k_1$. For the minimum principal curvature $k_2$ DeepFit outperforms all methods for low and no noise augmentation in addition to gradient and striped density augmentation,  however PCPNet has a small advantage for medium and high noise levels. The results for the minimum curvature are very sensitive since most values are close to zero. 
\begin{equation}
\label{eq:D_k}
    D_{k_j} = \left\lvert \frac{k_j-k_{GT}}{ \max\{|k_{GT}|, 1\}} \right\rvert, \quad \text{for } j=1,2.
\end{equation}

    \begin{table} 

	\parbox{.48\linewidth}{
		\centering	
		\begin{tabular}{| M{0.1\textwidth} | M{0.08\textwidth}| M{0.08\textwidth} | M{0.06\textwidth}|
		M{0.06\textwidth}|	M{0.06\textwidth}|} 
		\hline
		\centering\textbf{Aug.} &\textbf{Our DeepFit}  &
				\centering\textbf{PCP-Net} \cite{guerrero2018pcpnet} & \multicolumn{3}{c|}{\makecell[{{M{0.18\textwidth}}}]{\textbf{Jet} \\ \cite{cazals2005estimating}} }  	
		\tabularnewline
 			\hline
 			output & $k_1$+n &  $k_1$+n & $k_1$ & $k_1$ & $k_1$ \\ 
 			scale & ss & ms & small & med. & large \\
            \hlineB{2}
            None                & \textbf{1.00} & 1.36 &  2.19 &  6.55 & 2.97 \\
            \textbf{Noise $\sigma$}      &  & & & & \\
            $0.00125$  & \textbf{1.00} & 1.48 & 57.35 &  6.68 & 2.90 \\
            $0.006$    & \textbf{0.98} & 1.46 & 60.91 &  9.86 & 3.30 \\
            $0.012$    & \textbf{1.21}  & 1.59 & 49.40 & 10.78 & 3.58 \\
            \textbf{Density}    & & & & & \\
            Gradient            &\textbf{ 0.59 }& 1.32 &  2.07 &  1.40 & 1.53 \\
            Stripes             & \textbf{ 0.6 }& 1.09 &  2.04 &  1.54 & 1.89 \\
            \hline
		    \textbf{average}    & \textbf{0.89} & 1.38 & 28.99 &  6.13 &  2.69\\
		    \textbf{reduc.}    & \textbf{35.5\%} & &  &  & \\
		    \hline
		\end{tabular}
	}
\hfill
    \parbox{.48\linewidth}{
	\centering	
		\begin{tabular}{| M{0.1\textwidth} | M{0.08\textwidth}| M{0.08\textwidth} | M{0.06\textwidth}|
		M{0.06\textwidth}|	M{0.06\textwidth}|} 
		\hline
		\centering\textbf{Aug.} &\textbf{Our DeepFit}   &
				\centering\textbf{PCP-Net} \cite{guerrero2018pcpnet} & \multicolumn{3}{c|}{\makecell*[{{M{0.15\textwidth}}}]{\textbf{Jet} \\ \cite{cazals2005estimating} }}  	
		\tabularnewline
 			\hline
 			output & $k_2$+n  & $k_2$+n & $k_2$ & $k_2$ & $k_2$ \\ 
 			scale & ss & ms & small & med. & large \\
            \hlineB{2}
            None                & 0\textbf{.46} & 0.54 &  1.61 & 2.91 & 1.59  \\
            \textbf{Noise $\sigma$}      & & & & &  \\

            $0.00125$  & \textbf{0.47} & 0.53 & 25.83 & 2.98 & 1.53  \\
            $0.006$    & 0.57 & \textbf{0.51} & 22.27 & 4.88 & 1.73  \\
            $0.012$    & 0.68 & \textbf{ 0.53} & 18.17 & 5.22 & 1.84 \\
            \textbf{Density}    & & & & & \\
            Gradient            & \textbf{0.31} & 0.61 &  2.04 & 0.79 & 0.83 \\
            Stripes             & \textbf{0.31 }& 0.55 &  1.92 & 0.89 & 1.09 \\
            \hline
		    \textbf{average}    & \textbf{0.466} & 0.54 & 11.97 & 2.94 & 1.43\\
		    \textbf{reduc.}    & \textbf{13.7\%} & &  &  & \\
		    \hline
		\end{tabular}
	}
	\caption{Comparison of normalized RMSE for (left) maximal ($k_1$)  and (right) minimal ($k_2$) principal curvature  estimation of our DeepFit method to the classic Jet \cite{cazals2005estimating} with three scales, and PCPNet \cite{guerrero2018pcpnet}}
	\label{table:results:curvature_baselines}
\end{table}

\begin{figure}
\centering
	\includegraphics[width=0.98\linewidth]{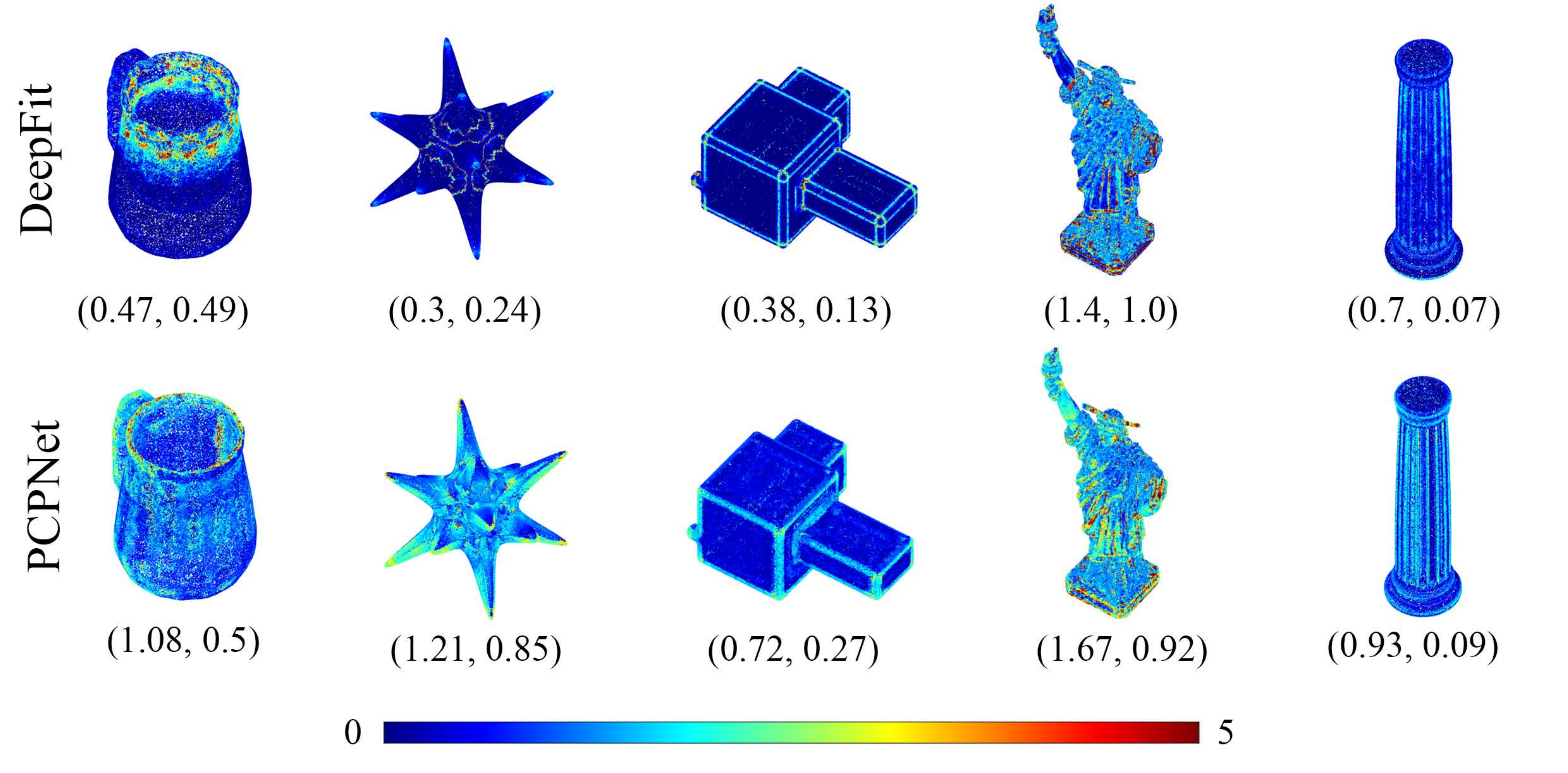}
	\caption{Curvature estimation error results for DeepFit compared PCPNet. The numbers under each point cloud are its normalized RMSE errors in the format ($k_1$, $k_2$). The color corresponds to the L2 norm of the error vector mapped to a heatmap ranging from 0-5.}
	\label{fig:results_curvature_error_comparison} 
\end{figure}

The normalized RMSE metric is visualized in Fig. \ref{fig:results_curvature_error_comparison} for DeepFit and PCPNet as the magnitude of the error vector mapped to a heatmap.  It can be seen that more errors occur near edges, corners and small regions with a lot of detail and high curvature.
These figures show that for both simple and complex geometric shapes DeepFit is able to predict the principal curvatures reliably. 

\subsection{Surface reconstruction and noise removal}
We further investigate the effectiveness of our surface fitting in the context of two subsequent applications---Poisson surface reconstruction~\cite{kazhdan2006poisson} and noise removal.

\subsubsection{Surface reconstruction.}
Fig. \ref{fig:results:poisson_recon} shows the results for the classical Jet fitting and our DeepFit approach. Since the reconstruction requires oriented normals, we orient the normals, in both methods, according to the ground truth normal. It shows that using DeepFit, the poisson reconstruction is moderately more satisfactory by being smoother overall, and crispier near corners. It also retains small details (liberty crown, cup rim).

\subsubsection{Noise removal.}
\begin{figure}[t]
\centering
    \begin{subfigure}{.48\textwidth}
        \centering
    	\includegraphics[width=0.98\linewidth]{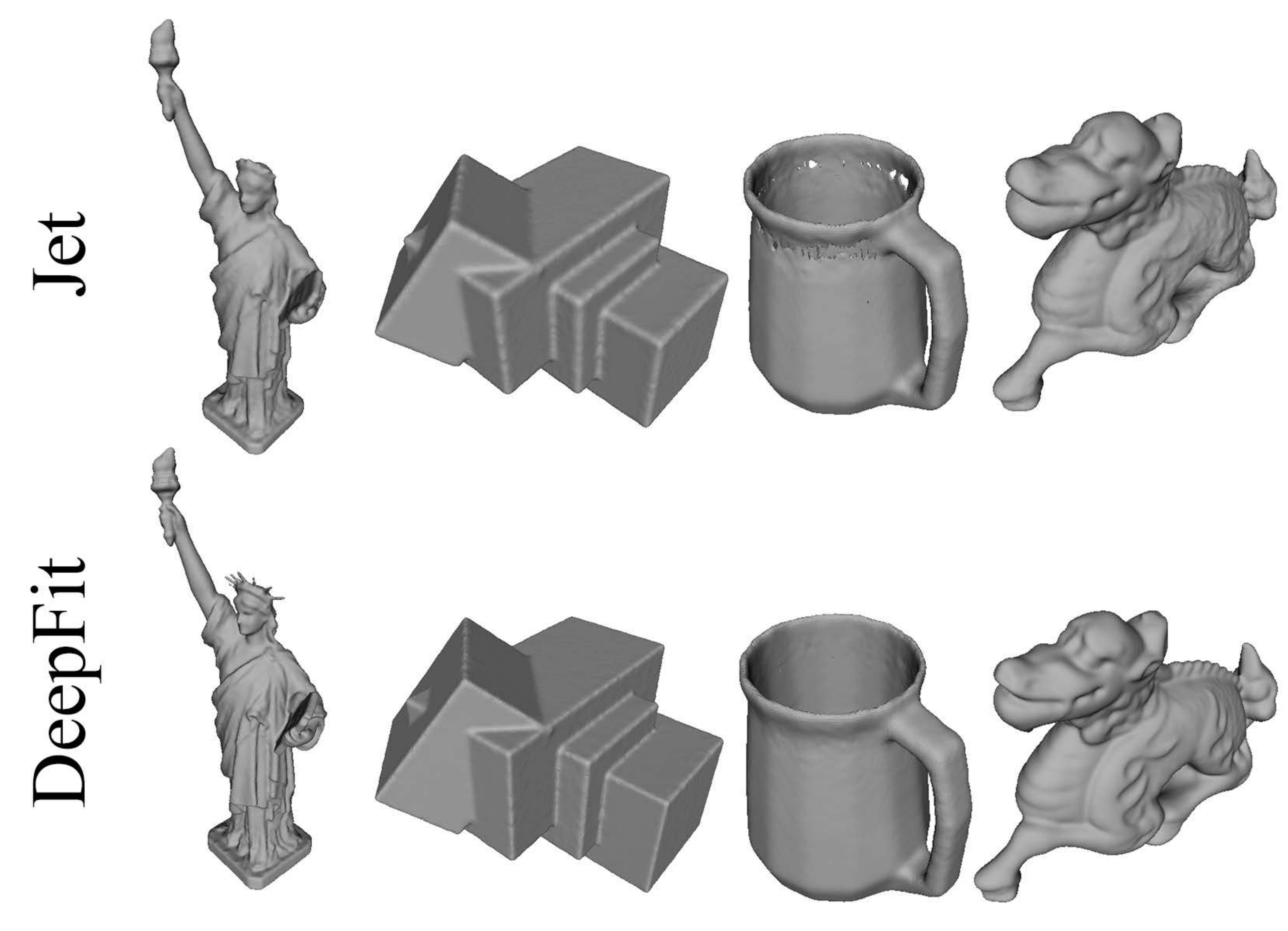}
    	\caption{}
    	\label{fig:results:poisson_recon} 
\end{subfigure}
    \unskip\ \vrule\ 
\begin{subfigure}{.48\textwidth}
    \centering
    \includegraphics[width=0.98\linewidth]{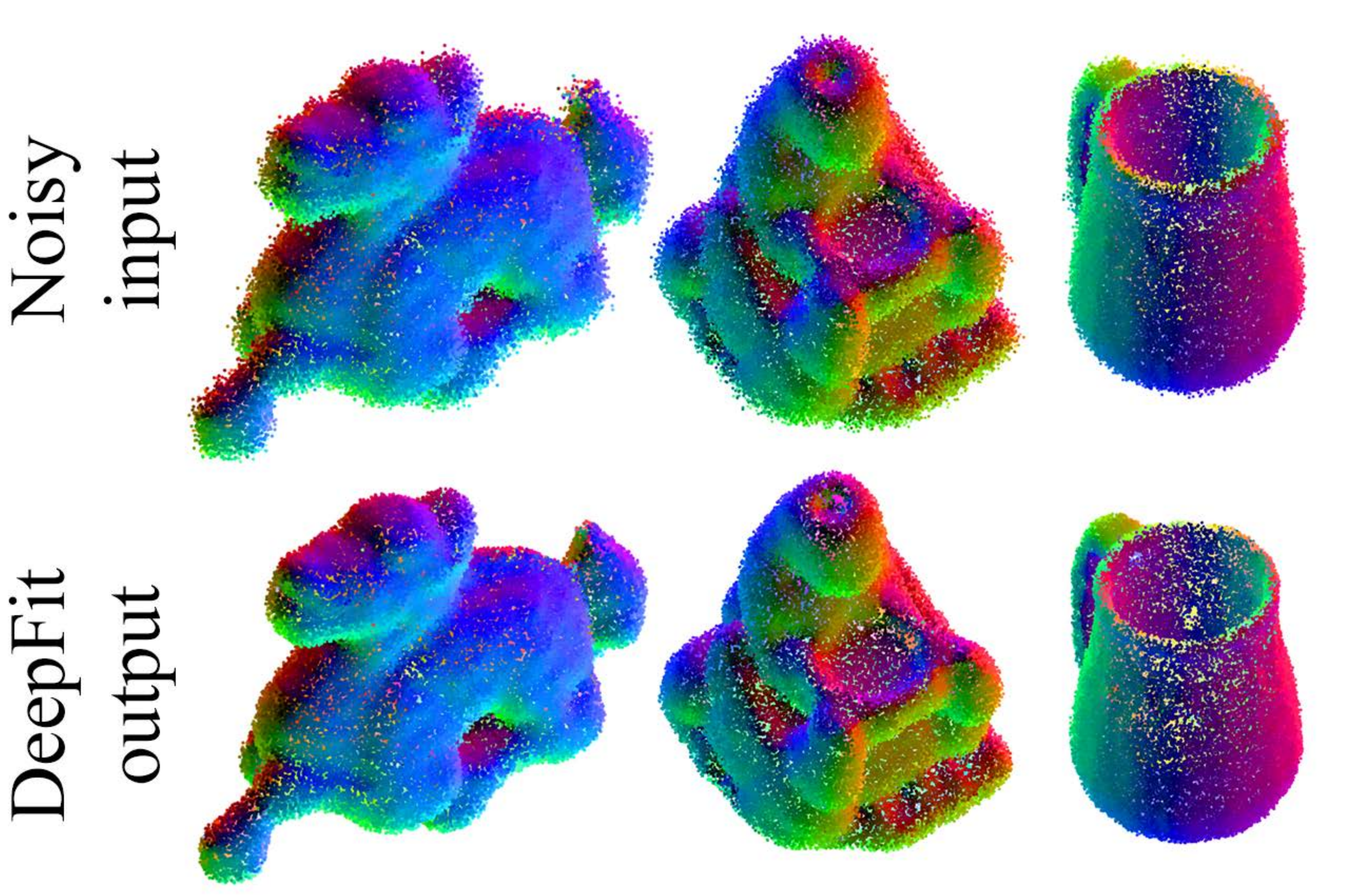}
    \caption{}
    \label{fig:results:noise_removal}
    \end{subfigure}
    \caption{DeepFit performance in two subsequent application pipelines: (a) Poisson surface reconstruction using estimated normal vectors from the classical Jet fitting and the proposed DeepFit. (b) Noise removal results using DeepFit predicted weights.}
\end{figure}

The point-wise weight prediction network enables a better fit by reducing the influence of neighboring points. This weight can also be interpreted as the network's confidence of that point to lie on the object's surface. Therefore, we can use the weight to remove points with low confidence.   We first aggregate the weights by summing all of its weight prediction from all of its neighbors. Then we compute the mean and standard deviation of the aggregateed weights and remove points under a threshold of $\mu(\sum w_i)-\sigma(\sum w_i)$.  The output point cloud contains less points than the original one and the removed points are mostly attributed to outliers or noise. The results are depicted in Fig. \ref{fig:results:noise_removal}.

\section{Summary }
\label{Sec:summary}

In this paper we presented a novel method for deep surface fitting for unstructured 3D point clouds. The method consists of estimating point-wise weights for solving a weighted least square fitting of an $n$-jet surface.
Our model is fully differentiable and can be trained end-to-end. The estimated weights (at test time) can be interpreted as the a confidence measure for every point in the point cloud and used for noise removal. Moreover, the formulation enables the computation of normal vectors and higher order geometric quantities like principal curvatures. The approach demonstrates high accuracy, robustness and efficiency compared to state-of-the-art methods. 
This is attributed to it’s ability to adaptively select the neighborhood of points through a learned model while leveraging classic robust surface fitting approaches, allowing the network to achieve high accuracy with a low number of parameters and computation time.

\bibliographystyle{plain}       
\bibliography{references}

\newpage
\section{Supplementary Material}
\label{Sec:appendix}

\subsection{Normal and principal curvature estimation performance}

\subsubsection{Performance on real data.} 
\begin{figure}
    \centering
    \includegraphics[width=0.95\linewidth]{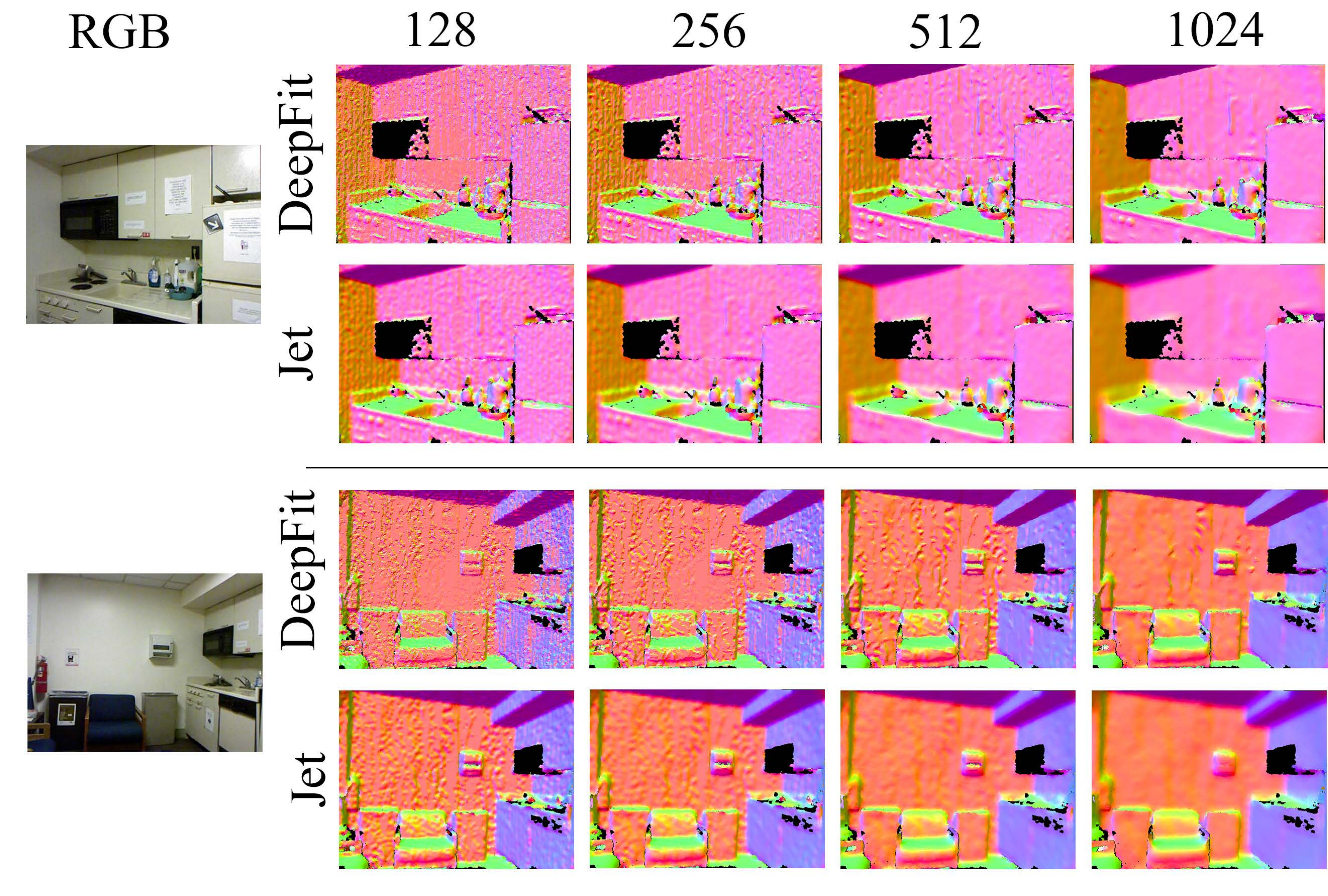}
    \caption{Normal estimation results for DeepFit and Jet on NYU Depth V2 dataset for different neighborhoods sizes ($128, 256, 512, 1024$). The colors of the points are normal vectors  mapped  to  RGB and projected to the image plane.}
    \label{fig:appx:results:nyu_v2}
    \vspace*{\floatsep}
    \centering
    \includegraphics[width=0.95\linewidth]{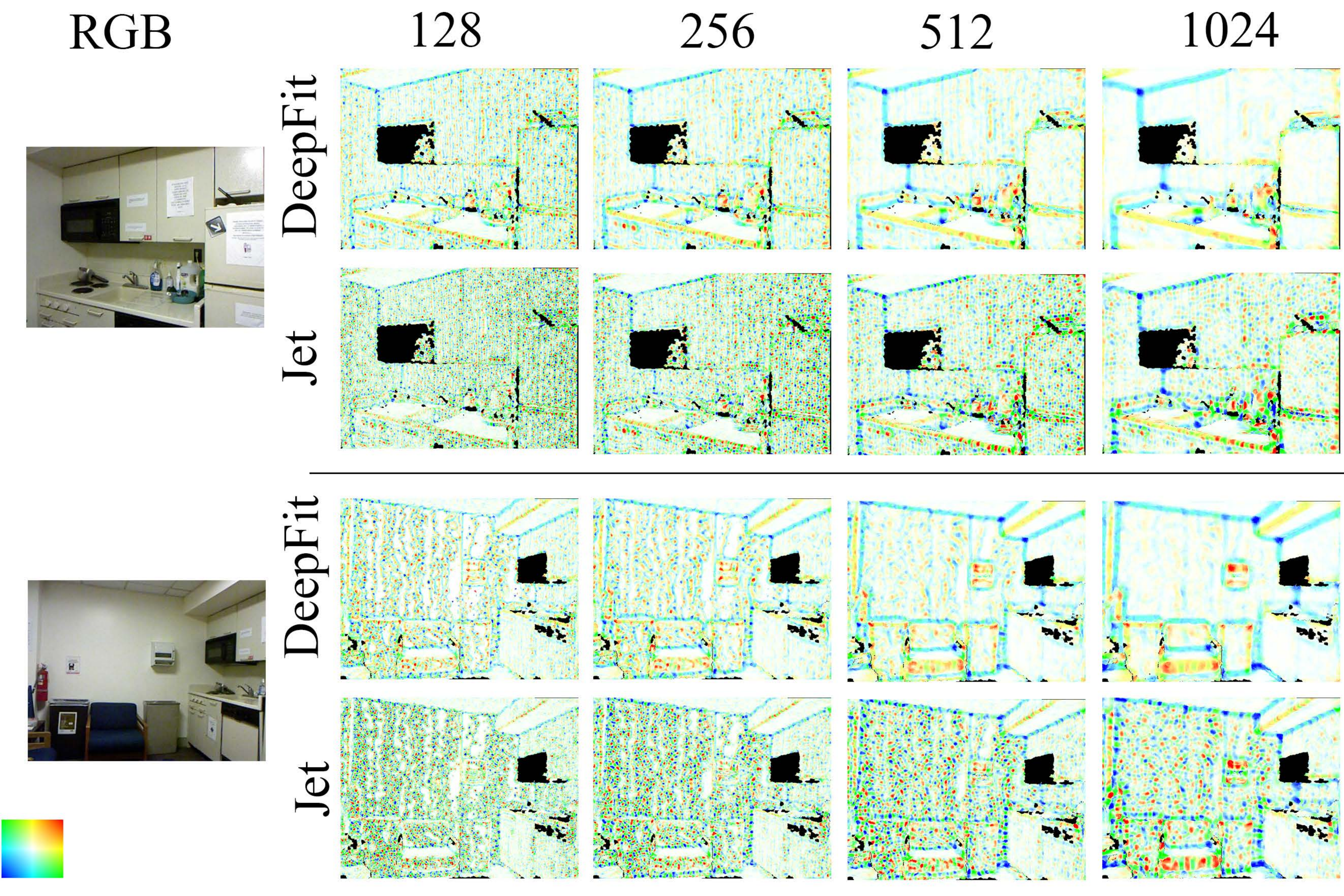}
    \caption{Principal curvature estimation results for DeepFit and Jet on NYU Depth V2 dataset for different neighborhoods sizes ($128, 256, 512, 1024$). The colors of the points correspond to their principal curvature values using the colormap in the bottom-left corner and projected to the image plane.}
    \label{fig:appx:results:nyu_v2_curv}
\end{figure}

We qualitatively evaluate the performance of our method on the NYU Depth V2 dataset \cite{SilbermanECCV12}. This dataset was captured using a Kinect v1 RGBD camera and contains indoor scene environment and includes missing data and a noise pattern that is significantly different than the PCPNet dataset. Specifically, the noise often has the same magnitude as some of the features. Most importantly, this dataset, much like other real-world datasets, does not have ground truth normals. Fig. \ref{fig:appx:results:nyu_v2} and Fig. \ref{fig:appx:results:nyu_v2_curv} show the performance of DeepFit's normal and principal curvature estimation respectively compared to Jet. DeepFit was trained with 256 points, however, since the network's weights are shared between the points it can be used with any neighborhood size. In these results we show the performance for $128, 256, 512, 1024$ neighboring points. It shows that DeepFit is less sensitive to noise and is able to overcome the over-smoothing affect commonly attributed to using a large neighborhood while also preserving fine details. 

\subsubsection{Additional normal estimation results.} 
\begin{figure}
\centering

	\begin{subfigure}{.32\textwidth}
    \centering
	    \includegraphics[width=0.98\linewidth]{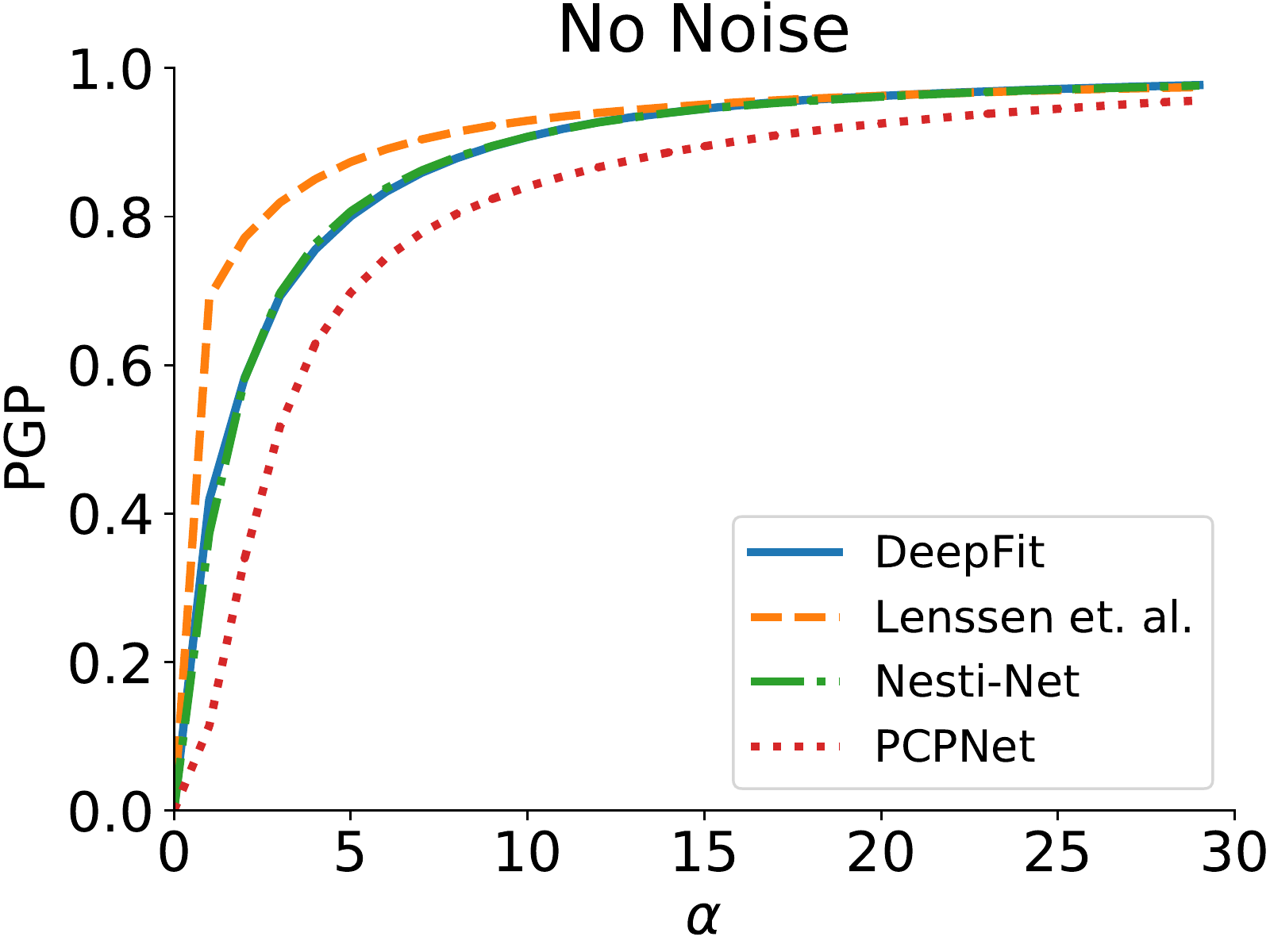}
  \label{fig:results:baselines_pgp:no_noise}
\end{subfigure}
	\begin{subfigure}{.32\textwidth}
    \centering
	    \includegraphics[width=0.98\linewidth]{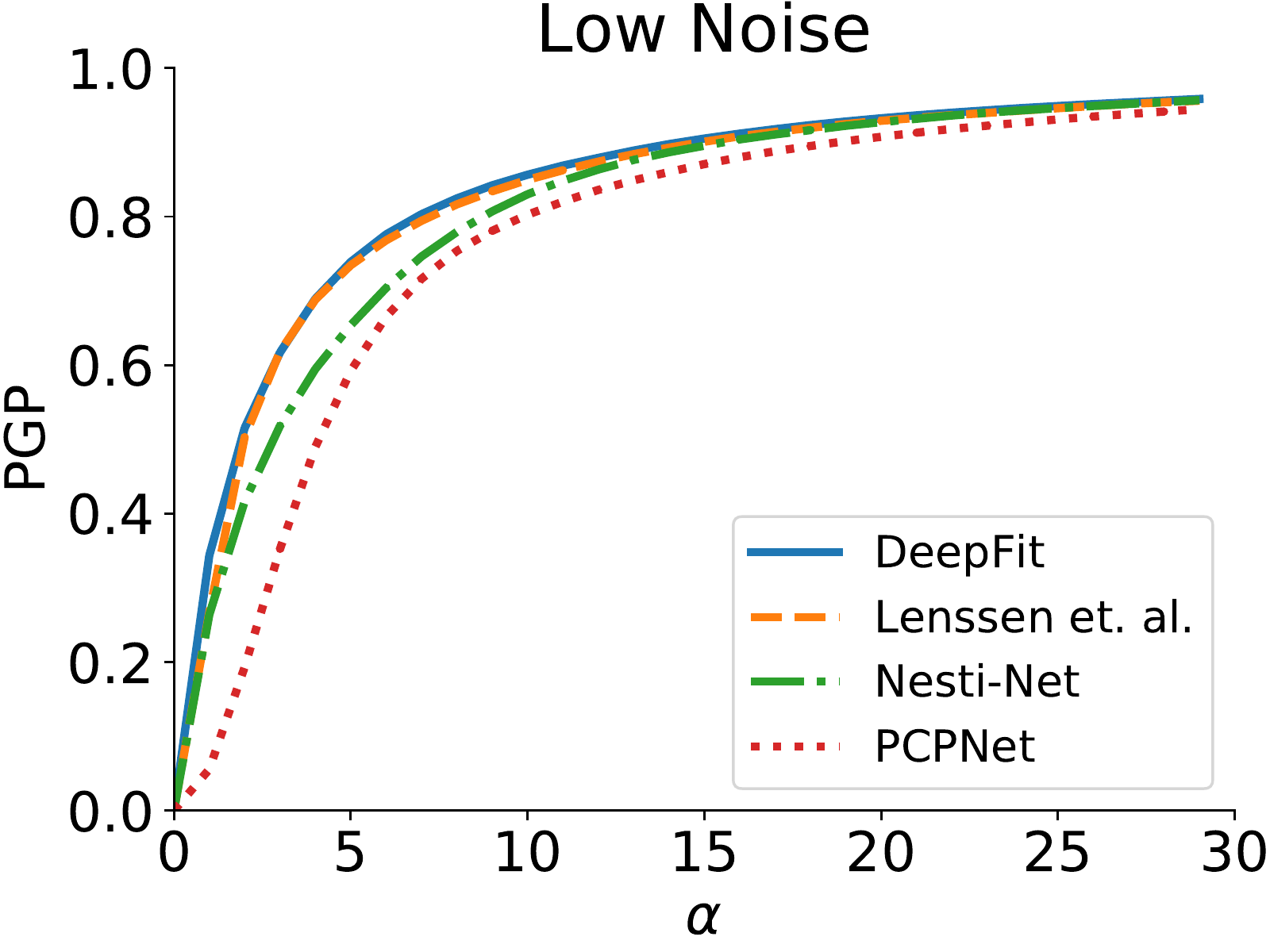}
  \label{fig:results:baselines_pgp:low_noise}
\end{subfigure}
	\begin{subfigure}{.32\textwidth}
    \centering
	    \includegraphics[width=0.98\linewidth]{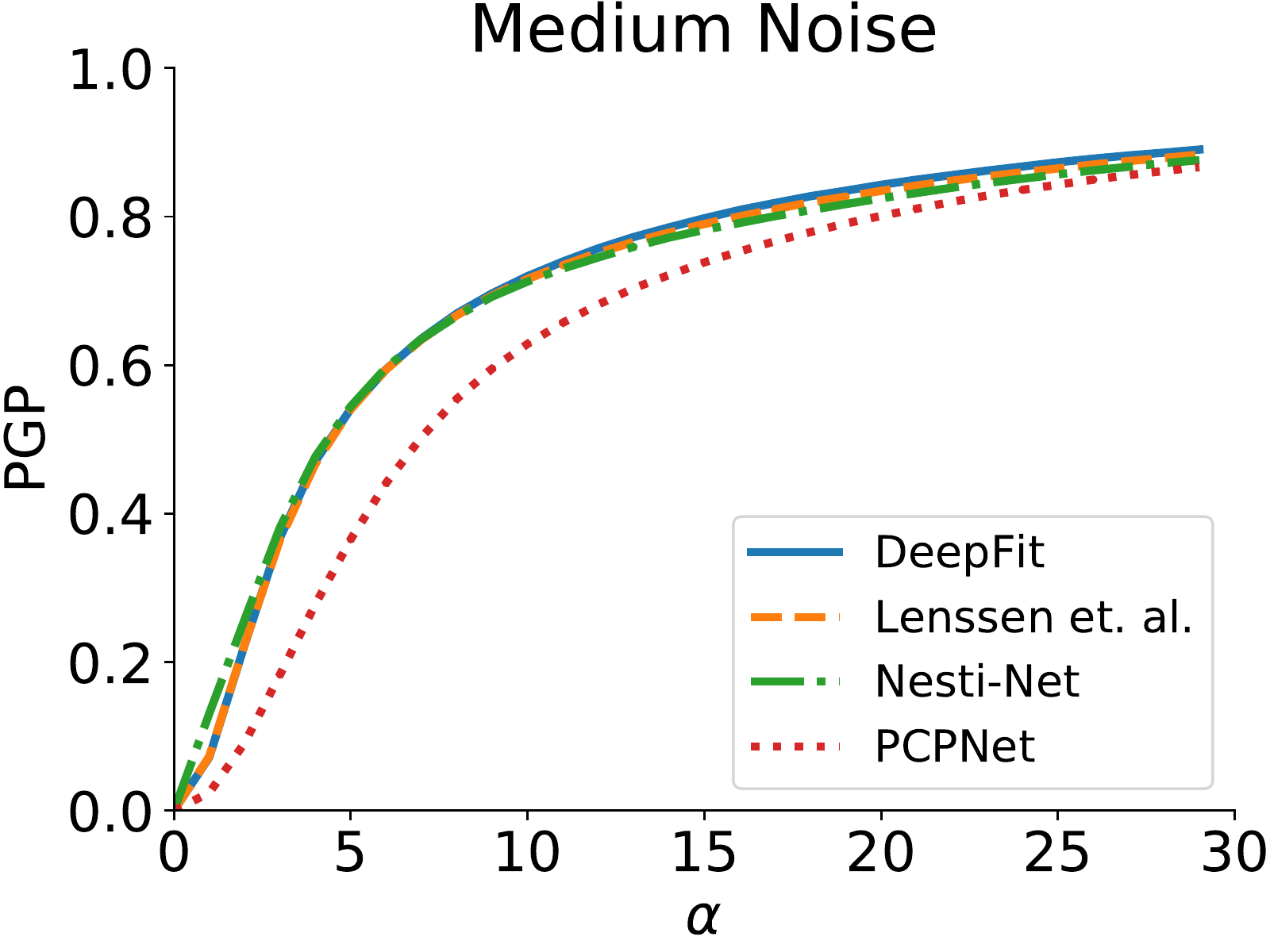}
  \label{fig:results:baselines_pgp:med_noise}
\end{subfigure}
	\begin{subfigure}{.32\textwidth}
    \centering
	    \includegraphics[width=0.98\linewidth]{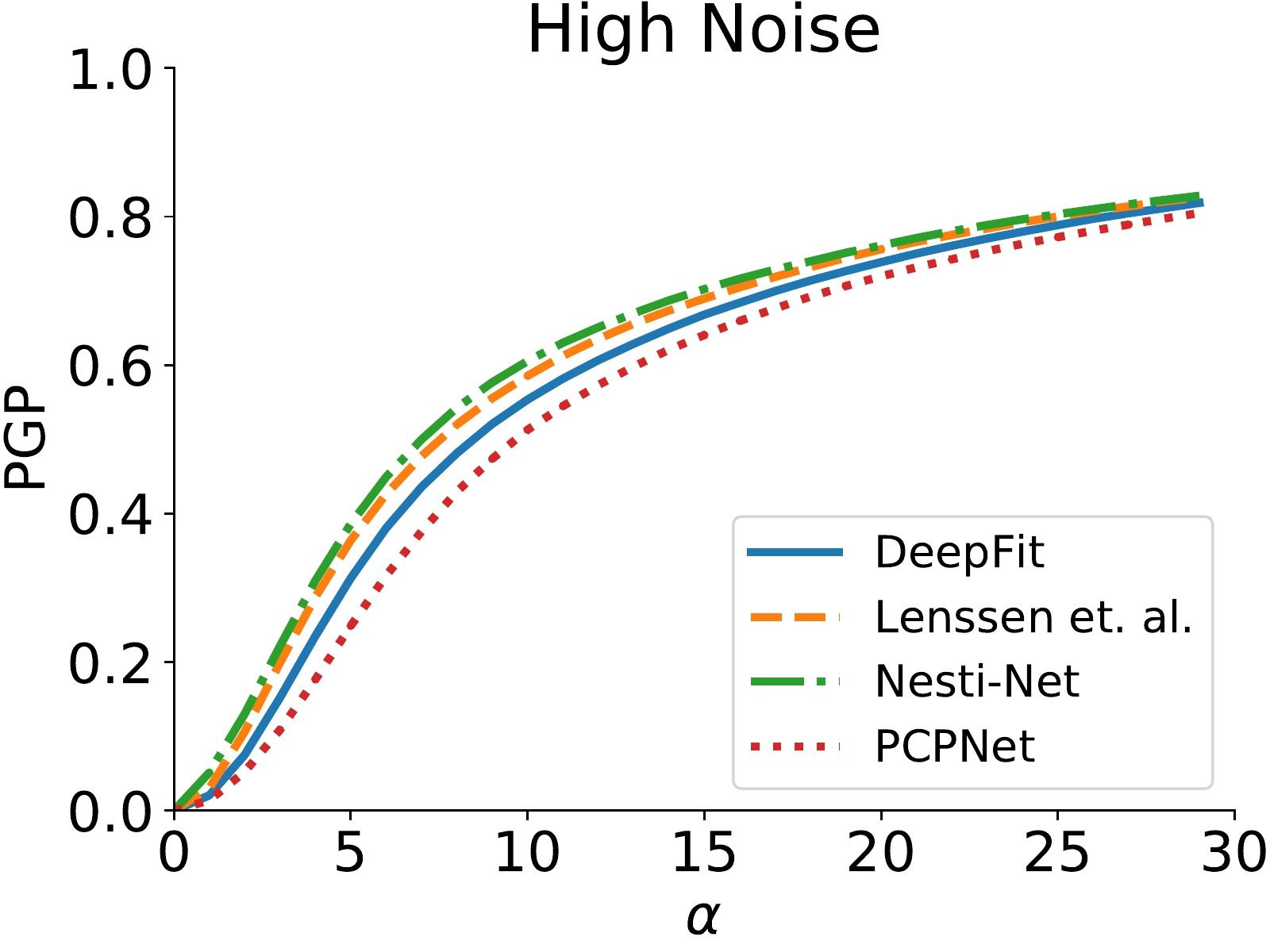}
  \label{fig:results:baselines_pgp:high_noise}
\end{subfigure}
	\begin{subfigure}{.32\textwidth}
    \centering
	    \includegraphics[width=0.98\linewidth]{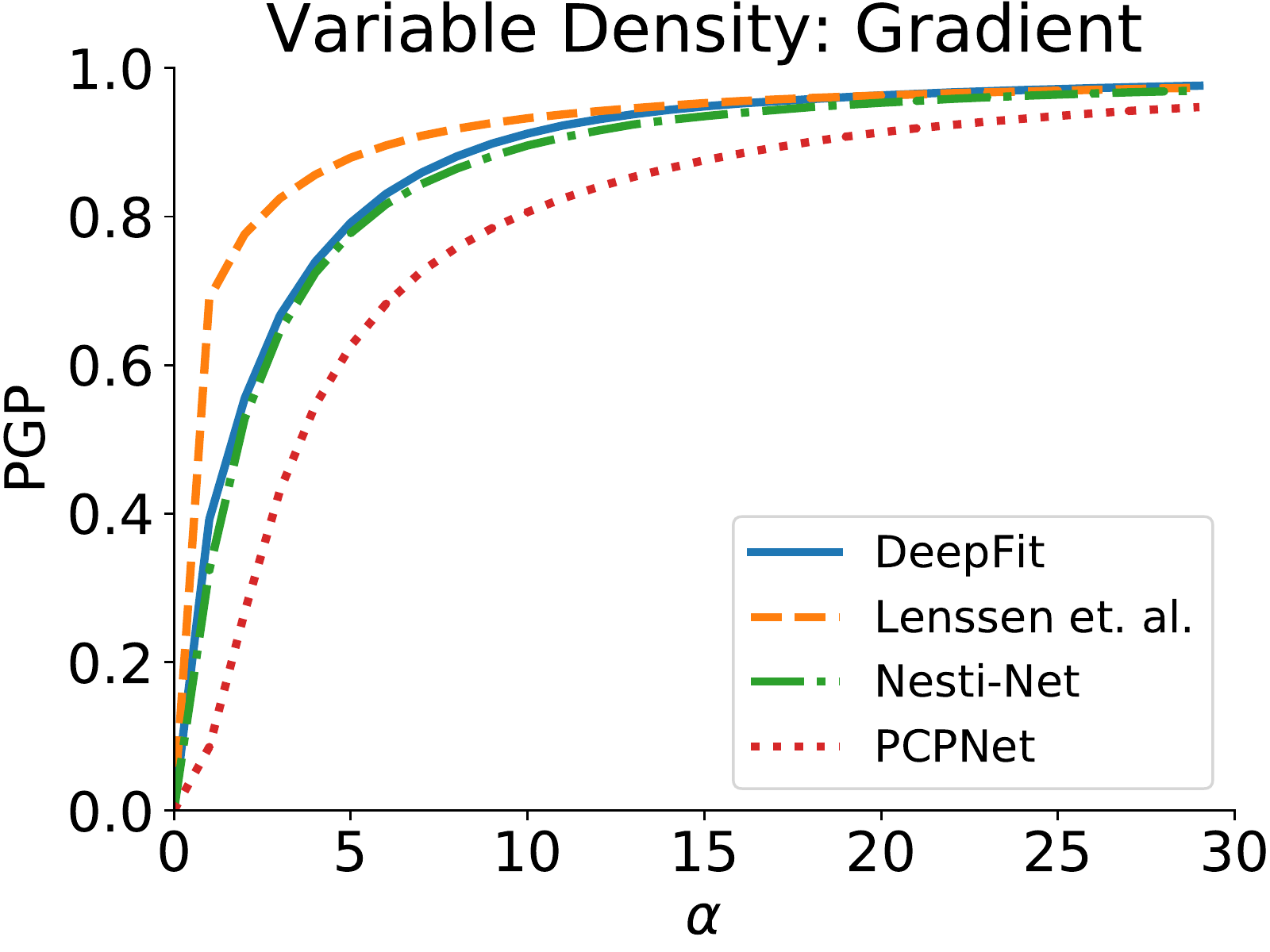}
  \label{fig:results:baselines_pgp:vardensity_grad}
\end{subfigure}
\begin{subfigure}{.32\textwidth}
    \centering
	    \includegraphics[width=0.98\linewidth]{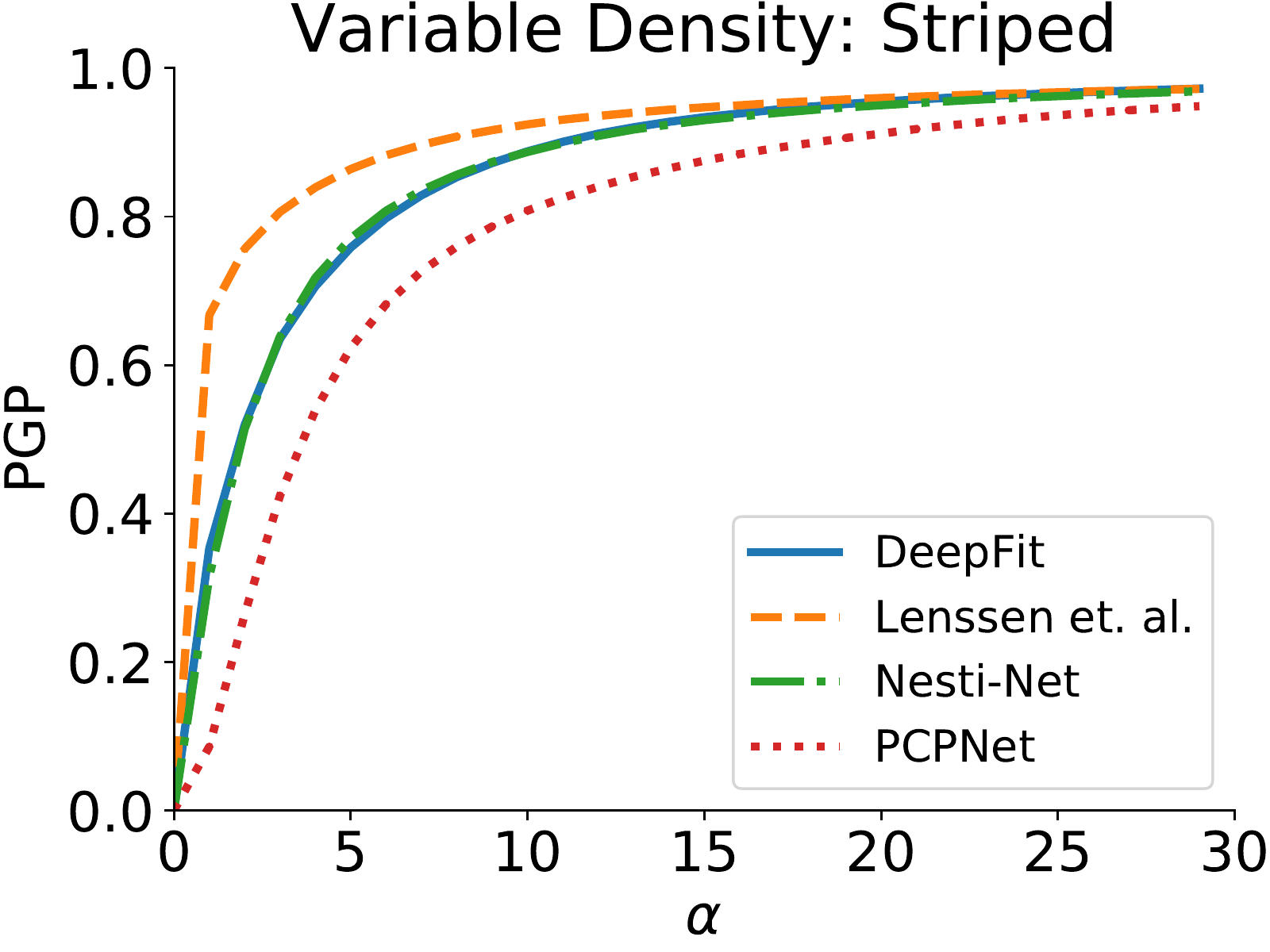}
  \label{fig:results:baselines_pgp:vardensity_stripe}
\end{subfigure}
	\caption{Comparison of the percentage of good points (PGP) metric for unoriented normal estimation of the proposed DeepFit to other deep learning methods (PCPNet \cite{guerrero2018pcpnet}, Nesti-Net \cite{ben2019nesti}, Lenssen et. al. \cite{lenssen2019differentiable}). Here, $\alpha$ is the threshold for measuring the percentage of good points.}.
	\label{figure:results:baselines_pgp}
\end{figure}

We evaluate the normal estimation performance on the PCPNet datast using the percentage of good points (PGP $\alpha$) metric. Fig \ref{figure:results:baselines_pgp} shows the results of different learning based methods for increasing $\alpha$ values . It shows that for low and medium noise levels, DeepFit is comparable to Lenssen et.al. \cite{lenssen2019differentiable} while in all other categories their performance is better. This is most likely attributed to the dataset bias towards flat and low curvature surfaces, in which case, our method does not pose an advantage. DeepFit main advantage is in curvy surfaces where an $n$-jet yields a better fit than a plane. 

\begin{figure}
\centering
    \begin{subfigure}{.98\textwidth}
    \centering
        \includegraphics[width=0.4\linewidth]{ablations_legend_rmse.pdf}
	\end{subfigure}
	\begin{subfigure}{.32\textwidth}
    \centering
	    \includegraphics[width=0.98\linewidth]{ablations_angle_rmse_results_no_noise.pdf}
  \label{fig:appx:results:baselines_rmse:no_noise}
\end{subfigure}
	\begin{subfigure}{.32\textwidth}
    \centering
	    \includegraphics[width=0.98\linewidth]{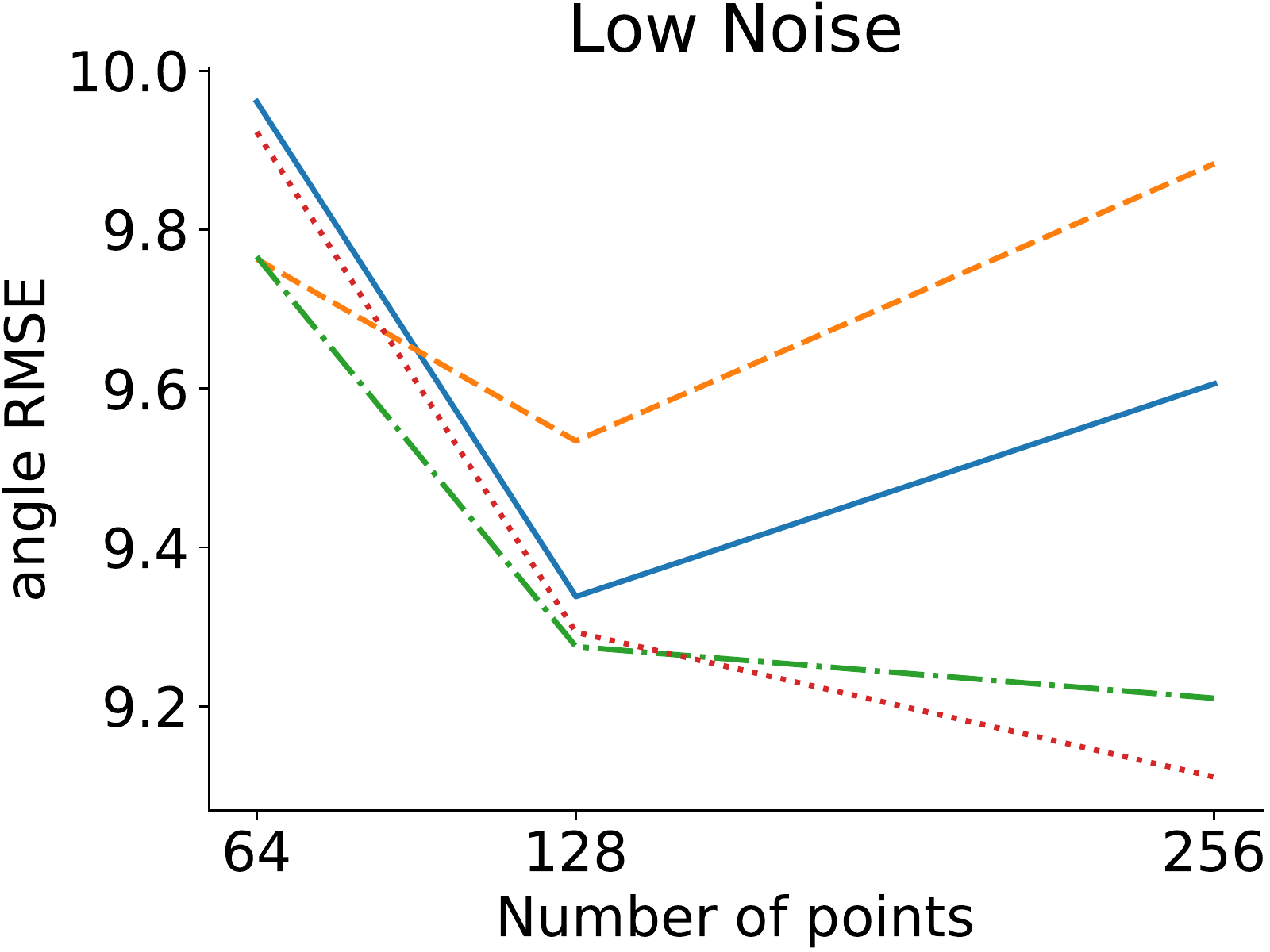}
  \label{fig:appx:results:baselines_rmse:low_noise}
\end{subfigure}
	\begin{subfigure}{.32\textwidth}
    \centering
	    \includegraphics[width=0.98\linewidth]{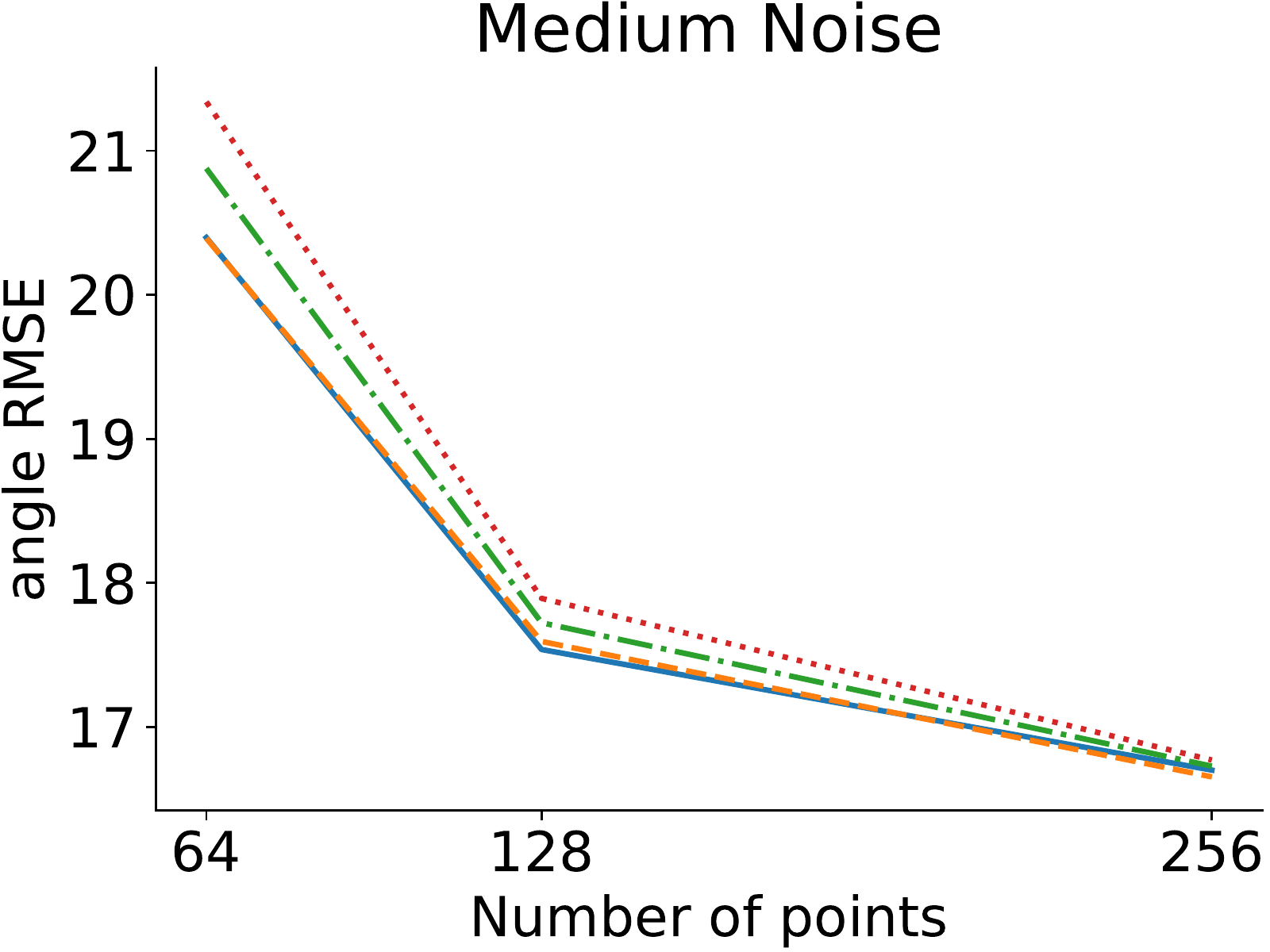}
  \label{fig:appx:results:baselines_rmse:med_noise}
\end{subfigure}
	\begin{subfigure}{.32\textwidth}
    \centering
	    \includegraphics[width=0.98\linewidth]{ablations_angle_rmse_results_high_noise.pdf}
  \label{fig:appx:results:baselines_rmse:high_noise}
\end{subfigure}
	\begin{subfigure}{.32\textwidth}
    \centering
	    \includegraphics[width=0.98\linewidth]{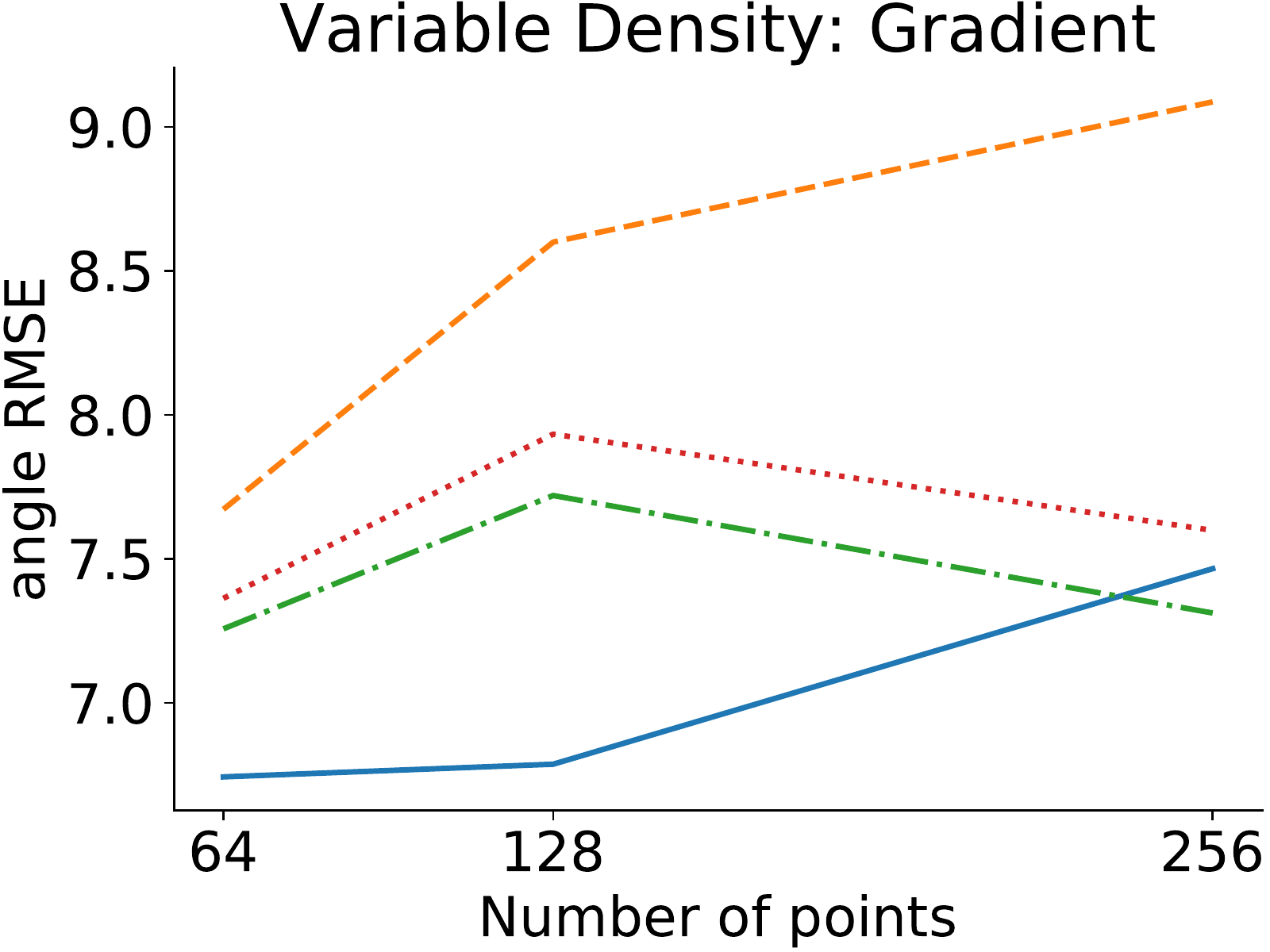}
  \label{fig:appx:results:baselines_rmse:vardensity_grad}
\end{subfigure}
\begin{subfigure}{.32\textwidth}
    \centering
	    \includegraphics[width=0.98\linewidth]{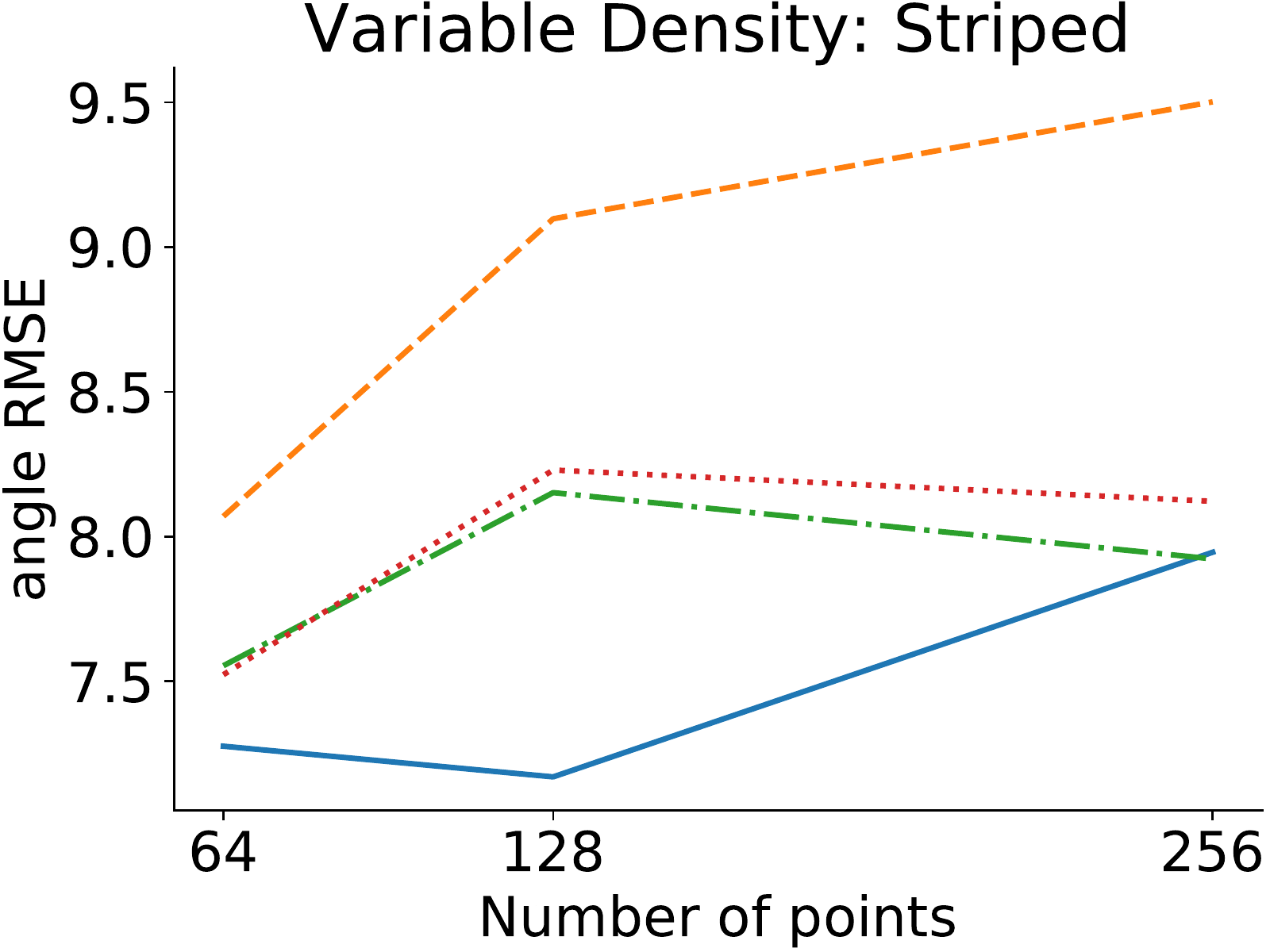}
  \label{fig:appx:results:baselines_rmse:vardensity_stripe}
\end{subfigure}
	\caption{Comparison of the angle RMSE metric for different DeepFit variants. Ablations include different $n$-jet order ($1, 2, 3, 4$) and number of neighboring points ($64, 128, 256$).}.
	\label{figure:appx:results:rmse_ablations_all}
\end{figure}

We evaluate DeepFit's normal estimation performance using RMSE for different $n$-jet orders and number of points in the neighborhood. The results are shown in Fig. \ref{figure:appx:results:rmse_ablations_all}. It shows that the increase in the number of neighboring points slightly decreases the performance in the no noise augmentation however it significantly improves the performance in high noise. This is mainly attributed to the weight estimation network that softly selects the most relevant points for the fit. It also shows that 1-jet (planes) perform well, however higher order jets have an advantage in the low and medium noise augmentation categories. In theory, the higher order jets have the capacity to fit planes, however in practice it is not always the case.

\begin{figure}
    \centering
    \includegraphics[width=0.98\linewidth]{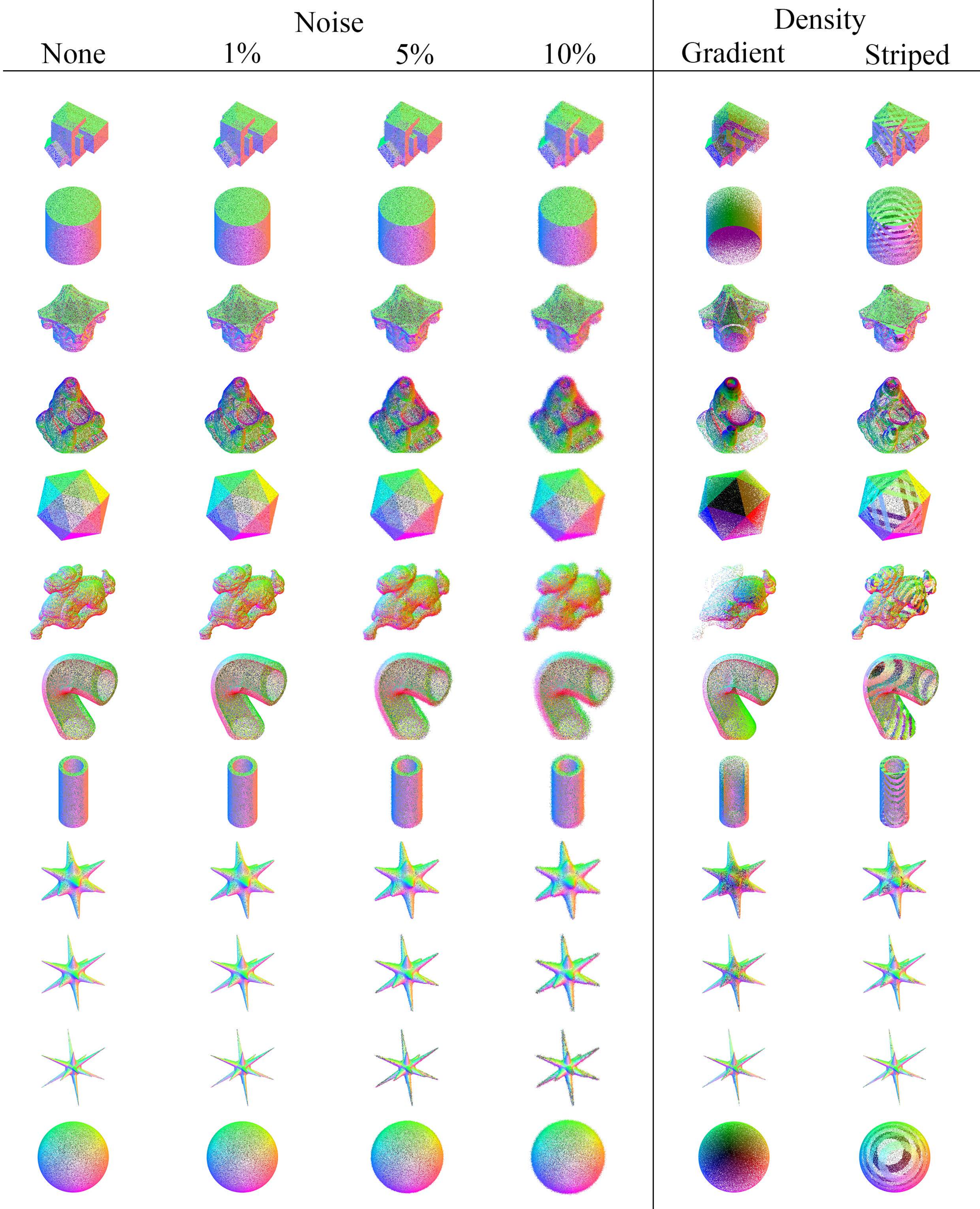}
    \caption{DeepFit's normal estimation results for different noise levels (columns 1-4), and density distortions (columns 5-6). The colors of the points are normal vectors mapped to RGB.}
    \label{fig:appx:results:pcpnet_all}
\end{figure}

\begin{figure}
    \centering
    \includegraphics[width=0.98\linewidth]{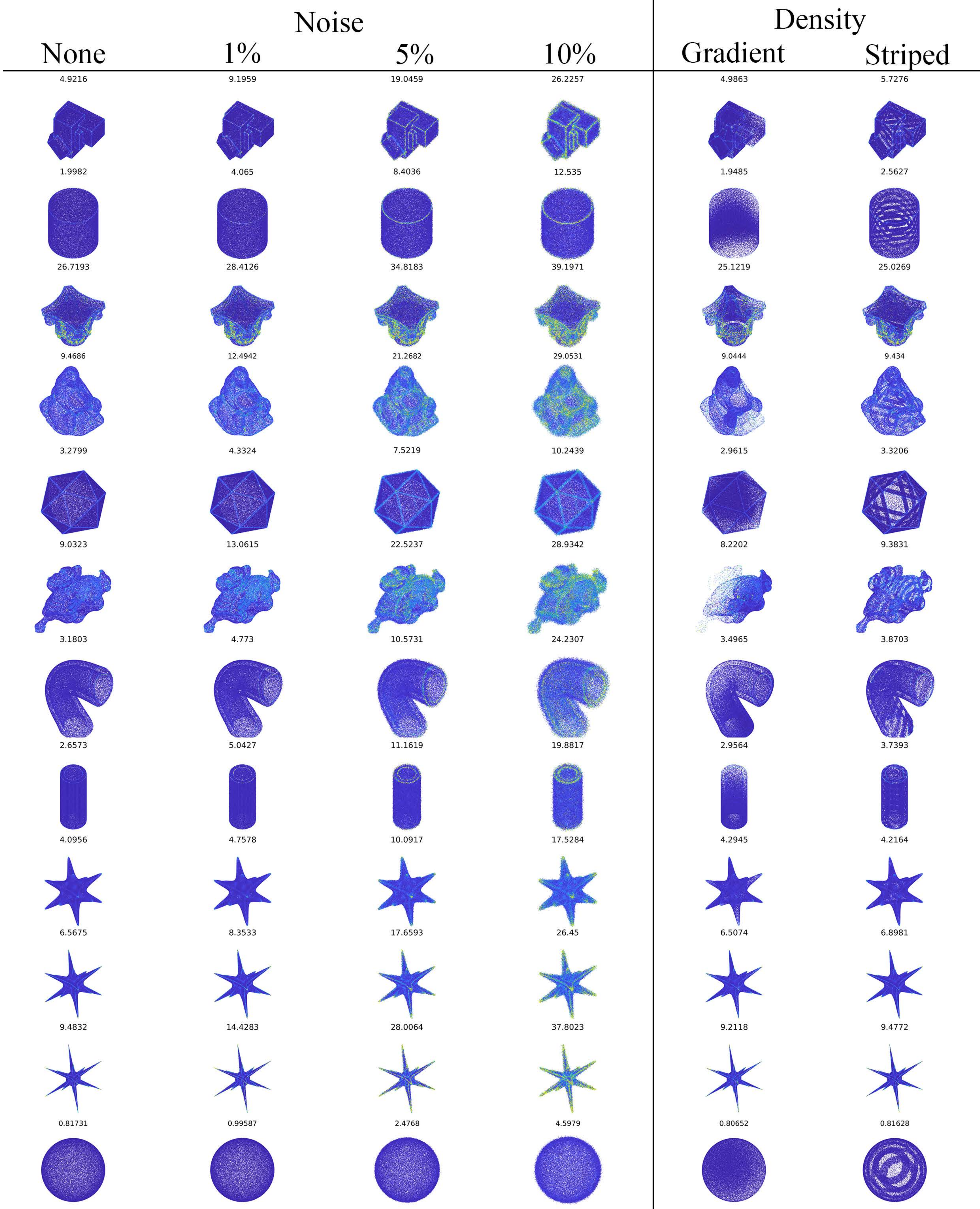}
    \caption{Normal estimation error visualization for different noise levels (columns 1-4), and density distortions (columns 5-6). The colors of the points correspond to angular difference, mapped to a heatmap ranging from 0-60 degrees. The number above each point cloud is the RMSE.}
    \label{fig:appx:results:pcpnet_all_err}
\end{figure}

Fig. \ref{fig:appx:results:pcpnet_all} depicts a visualization of DeepFit's results on PCPNet point clouds. Here the normal vectors are mapped to the RGB cube. 
Fig. \ref{fig:appx:results:pcpnet_all_err} depicts a visualization of the angular error in each point for the PCPNet dataset. Here,  the points' color correspond to angular difference, mapped to a heatmap ranging from 0-60 degrees. It shows that for complex shapes with high noise levels, the  general direction of the normal vector is predicted correctly, but, the fine details and exact normal vector are not obtained. For basic shapes the added noise does not affect the results substantially. Most notably, DeepFit shows robustness to point density corruptions. 
\subsubsection{Additional principal curvature estimation results.} 
\begin{figure}
    \centering
    \includegraphics[width=0.98\linewidth]{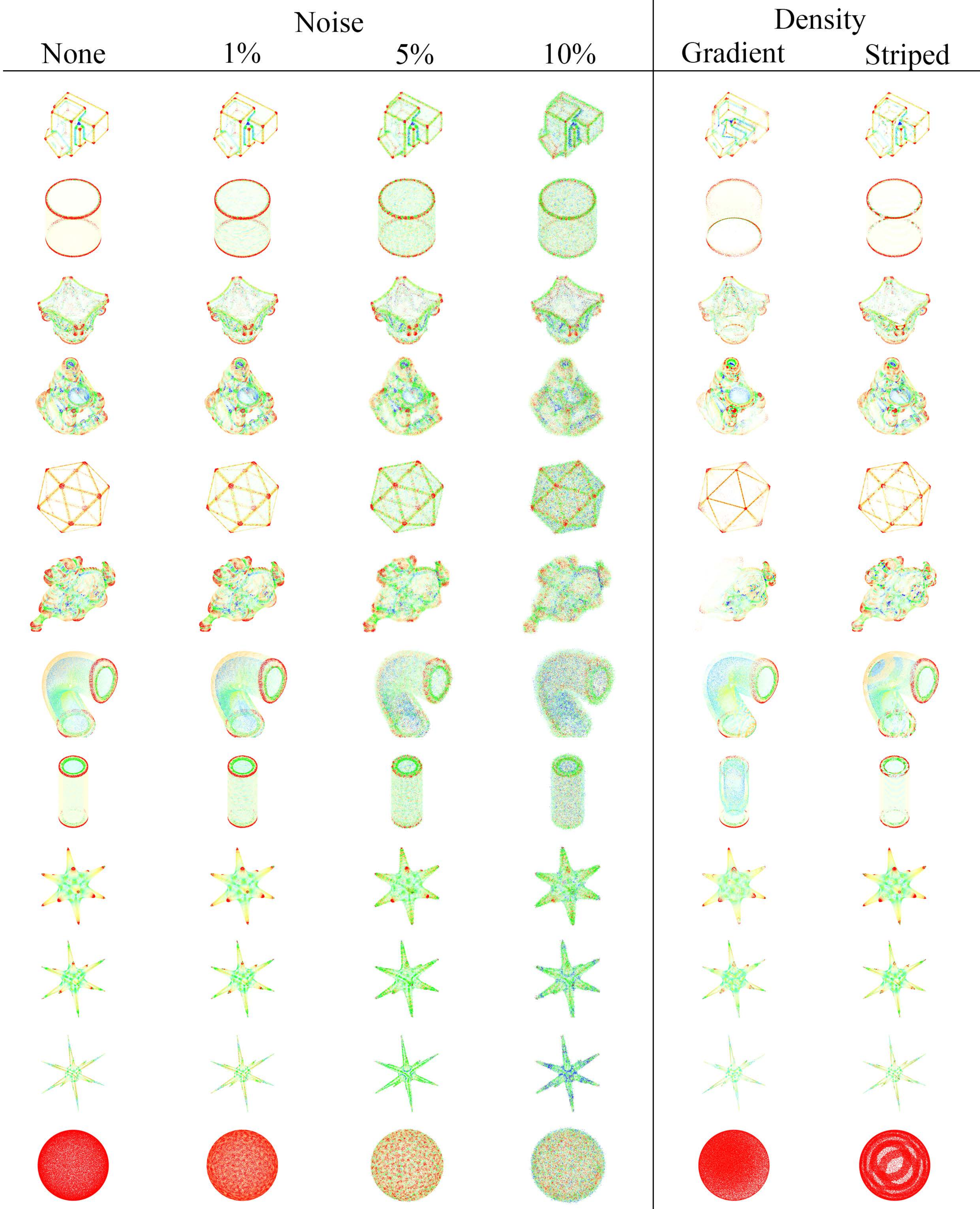}
    \caption{Curvature estimation results visualization.  The colors of the points corresponds to the mapping of $k_1, k_2$  to the color map given in Fig \ref{fig:curvature_result_visualization}. Values in the range $[-(\mu(|k_i|)+\sigma(|k_i|)),\mu(|k_i|)+\sigma(|k_i|)] |_{i=1,2}$.}
    \label{fig:appx:results:curv_pcpnet_all_err}
\end{figure}

\begin{figure}
    \centering
    \includegraphics[width=0.98\linewidth]{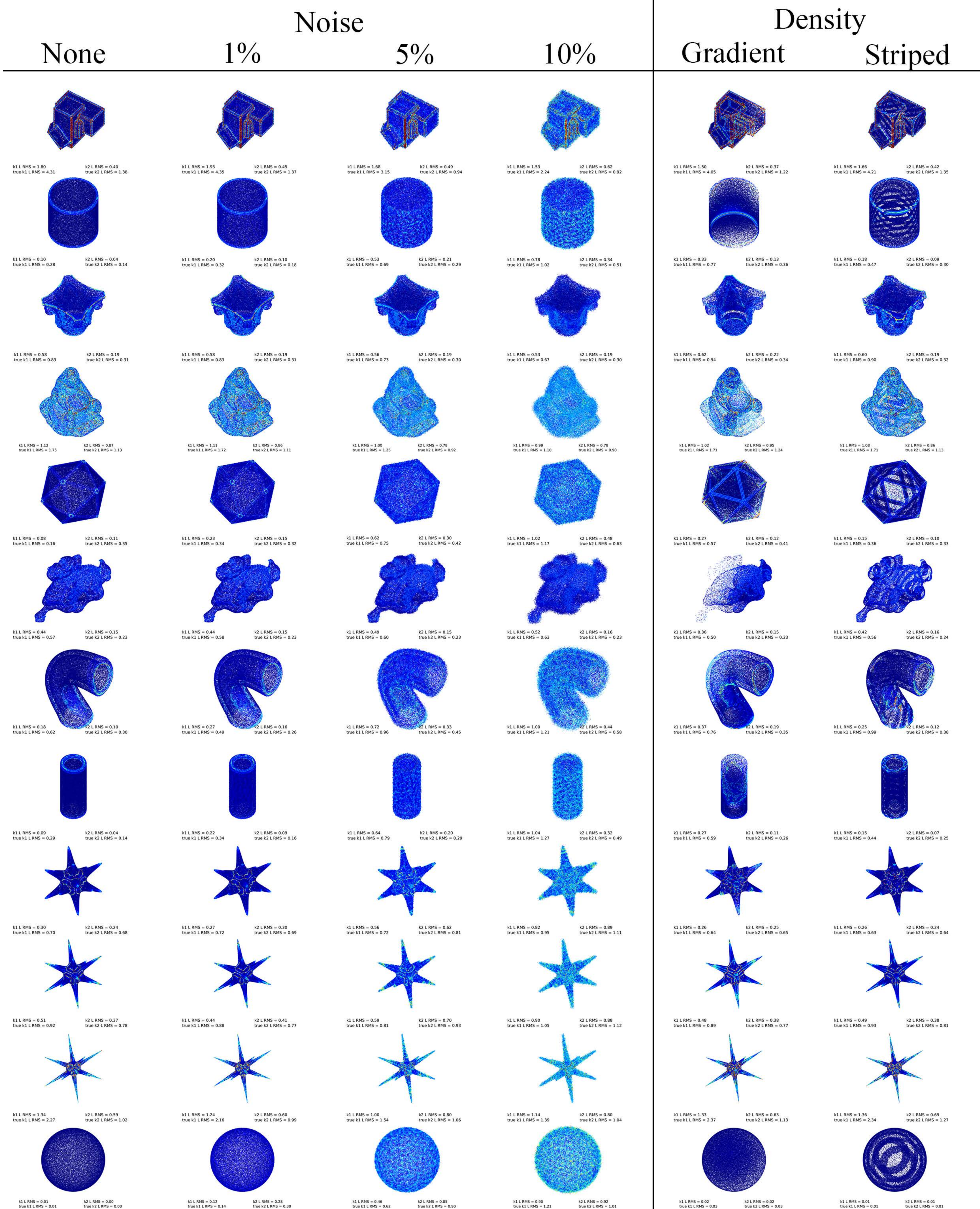}
    \caption{Curvature estimation error results. The numbers under each point cloud are its RMSE and normalized RMSE. The color corresponds to the L2 norm of the error vector mapped to a heatmap ranging from 0-5.}
    \label{fig:appx:results:curv_err_pcpnet_all_err}
\end{figure}

Fig. \ref{fig:appx:results:curv_pcpnet_all_err} qualitatively depicts DeepFit's results on the PCPNet dataset. For visualization, the principal curvatures are mapped to RGB values according to the commonly used mapping given in Fig. \ref{fig:curvature_result_visualization} i.e. both positive (dome) are red, both negative (bowl) are blue, one positive and one negative (saddle) are green, both zero (plane) are white, and one zero and one positive/negative (cylinder) are yellow/cyan. For consistency in color saturation we map each model differently according to the mean and standard deviation of the principal curvatures.  Note that the curvature sign is determined by the ground truth normal orientation. 
DeepFit's normalized RMSE metric is visualized in Fig. \ref{fig:appx:results:curv_err_pcpnet_all_err}  as the magnitude of the error vector mapped to a heatmap.  It can be seen that more errors occur near edges, corners and small regions with a lot of detail and high curvature.
Moreover, these visualizations show that for low noise levels, the principal curvature estimation is reliable, as expected, the reliability declines with the insertion of high magnitude noise.  

\subsection{Efficiency}
Table \ref{table:parameter_comparisong} shows a comparison between the number of parameters and run time between different deep learning based normal estimation methods. It can be seen that DeepFit has a significantly lower number of parameters compared to Nesti-Net and PCPNet and more parameters than Lenssen et. al.. This gap in the number of parameters can be explained by the lack of point structure in our method while Lessen et. al. construct a graph. Constructing a graph introduces a limitation with respects to the number of neighboring points, i.e. training and testing has to be done on the same neighborhood size, using a PointNet architecture allows to train and test on different sizes of neighborhoods.  DeepFit's , number of parameters is mainly attributed to the PointNet transformation sub-networks. The reported run time is the average run time for a batch of size 64 (i.e. computing normals for 64 points simultaneously). We chose a batch of 64 in order to fairly compare to the more resource intensive methods (Nesti-Net). Most methods, including ours, can compute in larger batches for faster performance, particularly Lenssen et. al. that are able to fit a full size point cloud (100k points) in a single batch on the GPU. 

\begin{table}[]
    \centering
    \begin{tabular}{M{0.2\textwidth}  M{0.15\textwidth}  M{0.15\textwidth}  M{0.15\textwidth}  M{0.15\textwidth}}
    \toprule
        Method & Our DeepFit & Nesti-Net \cite{ben2019nesti}& PCPNet \cite{guerrero2018pcpnet}& Lenssen et. al. \cite{lenssen2019differentiable}\\
        \hline
        Parameters & 3.5M&179M& 22M& 7981\\
        Exec time (per point) & 0.35ms & 266ms & 0.61ms & 0.13ms\\
         \bottomrule
    \end{tabular}
    \caption{Number of parameters and execution time performance for deep learning normals estimation methods. Run time is averaged for batches of size 64. }
    \label{table:parameter_comparisong}
\end{table}

\end{document}